\documentclass[10pt,journal,compsoc]{IEEEtran}

\usepackage{amsmath,amsfonts}
\usepackage{url}
\usepackage{color}
\usepackage{booktabs}       
\usepackage{multirow}
\usepackage{multicol}
\usepackage{epsfig}
\usepackage{subcaption}
\usepackage{ragged2e}
\usepackage{amssymb}
\usepackage{utfsym}
\usepackage[ruled,linesnumbered]{algorithm2e}
\usepackage[pagebackref=true,breaklinks=true,colorlinks,bookmarks=false]{hyperref}

\definecolor{MocoGreen}{RGB}{57,181,74}
\newcommand{\etal}{\textit{et al.}}
\begin{document}

\title{Adversarial Robust Memory-Based \\
Continual Learner}
\author{Xiaoyue~Mi,
        Fan~Tang,
        Zonghan~Yang,
        Danding~Wang,
        Juan~Cao,
        Peng~Li,
        Yang~Liu 
\IEEEcompsocitemizethanks{
\IEEEcompsocthanksitem{Xiaoyue~Mi, Fan~Tang, Danding~Wang, and Juan~Cao are with the Institute of Computing Technology, Chinese Academy of Sciences, Beijing 100190,
China, e-mail: \{mixiaoyue19s,tangfan,wangdanding,caojuan\}@ict.ac.cn.} 
\IEEEcompsocthanksitem{Peng Li is with the Institute for AI Industry Research (AIR), Tsinghua University, Beijing 100084, China, e-mail: lipeng@air.tsinghua.edu.cn.}
\IEEEcompsocthanksitem{Zonghan~Yang and Yang Liu is with the Department of Computer Science and Technology, Tsinghua University, Beijing 100084, China, e-mail: yangzh20@mails.tsinghua.edu.cn,liuyang2011@tsinghua.edu.cn.}
}
}




\IEEEtitleabstractindextext{
\justify
\begin{abstract}
Despite the remarkable advances that have been made in continual learning, the adversarial vulnerability of such methods has not been fully discussed.
We delve into the adversarial robustness of memory-based continual learning algorithms and observe limited robustness improvement by directly applying adversarial training techniques.
Preliminary studies reveal the twin challenges for building adversarial robust continual learners: \textbf{accelerated forgetting} in continual learning and \textbf{gradient obfuscation} in adversarial robustness.
In this study, we put forward a novel adversarial robust memory-based continual learner that adjusts data logits to mitigate the forgetting of pasts caused by adversarial samples.
Furthermore, we devise a gradient-based data selection mechanism to overcome the gradient obfuscation caused by limited stored data.
The proposed approach can widely integrate with existing memory-based continual learning as well as adversarial training algorithms in a plug-and-play way.
Extensive experiments on Split-CIFAR10/100 and Split-Tiny-ImageNet demonstrate the effectiveness of our approach, achieving up to 8.13\% higher accuracy for adversarial data.

\end{abstract}


}

\maketitle


\section{Introduction}
\label{sec:intro}

\IEEEPARstart{C}{ontinual} learning~\cite{parisi2019continual} enables intelligent models to continually acquire new information, update their knowledge, and adapt to the evolving dynamics of their environment.
It has recently emerged as a promising direction to achieve ideal intelligent models, and some pioneer works have investigated the robustness of continual learning in various aspects~\cite{sarfraz2023error,wu2021pretrained,yoon2019scalable}.
Achieving robust continual learners represents a crucial advancement in enabling intelligent models to adapt to dynamic scenarios effectively.

However, when dealing with streaming non-\textit{i.i.d.} data, continual learners usually are problematic and vulnerable to adversarial examples~\cite{khan2022adversarially}.
Chen~\etal~\cite{chen2022queried} first attempt at preserving adversarial robustness in continual learning~\cite{li2017learning} by combining it with adversarial training~\cite{zhang2019theoretically} with the help of a large amount of unlabeled data.
Despite the impressive results of Chen~\etal~\cite{chen2022queried}, the inherent challenges for conducting adversarial robust continual learners are still under disclosure.

\begin{figure}[t]
  \centering
  \includegraphics[width=0.9\linewidth]{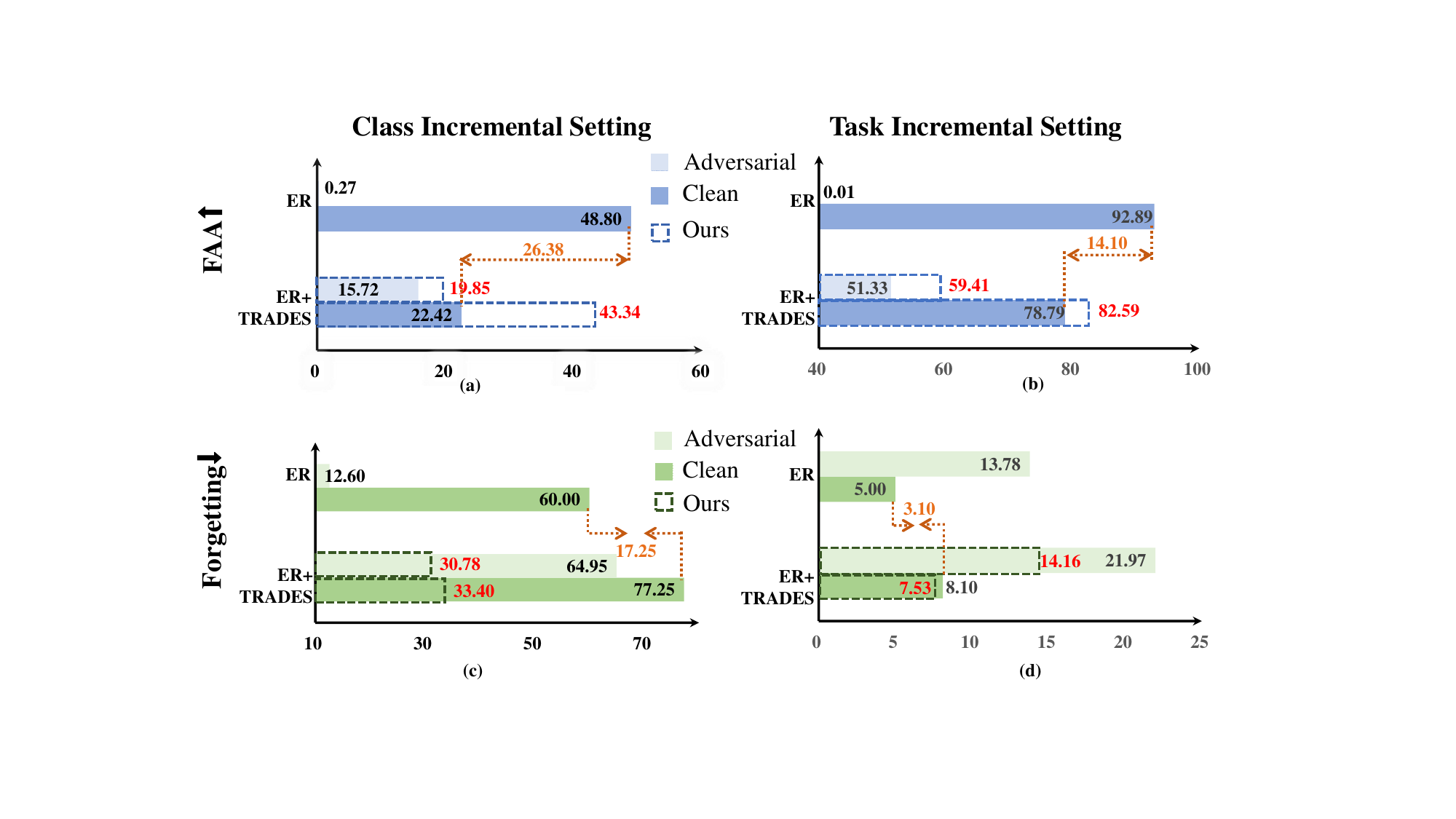}
  \caption{A preliminary analysis of the combination of continual learning and adversarial training on Split-CIFAR10 with buffer size $200$.
We consider the \textcolor[RGB]{68,114,196}{final average accuracy (FAA)} and \textcolor[RGB]{161,196,114}{forgetting} of adversarial and clean samples in both class incremental and task incremental settings.
The \textcolor[RGB]{202,12,22}{red numbers} indicate the performance after adding our approach.
}
  \label{fig:intro}
  \vspace{-10pt}
\end{figure}

In this study, we propose to undertake a deeper investigation into building adversarial robust continual learners without abundant unlabeled data.
We separately select two representative approaches for continual learning and adversarial defenses: memory-based continual learning and adversarial training.
Fig.~\ref{fig:intro} shows the results of a direct combination of memory-based continual learning (ER~\cite{riemer2019learning}) with an adversarial training approach (TRADES~\cite{zhang2019theoretically}).
The results for adversarial data reconfirm that continual learners risk adversarial attacks~\cite{khan2022adversarially,chen2022queried,guoattacking,khan2022susceptibility} and could be strengthened by adversarial training~\cite{chen2022queried}.
Whereas, the increase in \textcolor[RGB]{161,196,114}{forgetting} for clean data shows that combining with adversarial training could weaken the performance of continual learning.

Further experimental analysis in Sec.~\ref{sec:analysis} reveals that adversarial training and continual learning interfere with each other from two aspects: (1) adversarial training accelerates the forgetting of past tasks or classes in continual learning due to the increased negative gradients toward past classes caused by adversarial examples;
(2) continual learning exacerbates gradient obfuscation~\cite{athalye2018obfuscated} in adversarial training, which is a phenomenon of biased gradient information compared with joint training caused by insufficient historical data.
These two challenges arise naturally with the simple combination of continual learning and adversarial training.

Based on the analysis, we propose an adversarial robust memory-based continual learner without a large amount of additional data, which consists of an anti-forgettable logit calibration (AFLC) module and a robustness-aware experience replay (RAER) strategy.
The AFLC module designs different logit calibration strategies for past, current, and future classes, thus mitigating the negative gradients of adversarial samples for continual learning. 
By leveraging attacking difficulty, the RAER strategy selects data points that are adversarial safety and maintains the diversity of data distribution, rather than storing data that overfitting to the decision boundary~\cite{chaudhry2018riemannian}.
Our approach can be combined with most memory-based continual learning as well as adversarial training algorithms, and we have demonstrated the effectiveness of our approach on the Split-CIFAR10/100 and the more challenging Split-Tiny-ImageNet datasets.

Our contributions can be summarized as follows:
\begin{itemize}
\item We reveal the twin challenges to carrying out memory-based continual learning algorithms: accelerated forgetting for continual learning and gradient obfuscation for adversarial robustness.
    \item We propose anti-forgettable logit calibration by adapting the logits of past, current, and future tasks to reduce negative influences from adversarial training and robustness-aware experience replay by storing robust and diverse past data to alleviate gradient obfuscation.
    \item Experiments on Split-CIFAR10/100 and Split-Tiny-ImageNet demonstrate our effectiveness in improving both clean accuracy and adversarial robustness of continual learning and mitigating forgetting in both class-incremental and task-incremental settings without additional data.
\end{itemize}

\section{Related Works}
\label{sec:related_work}
\noindent \textbf{Continual learning.}
Continual learning~\cite{parisi2019continual} tries to make the model adapted to the changed data distribution following time while containing the knowledge of past data.
For the main challenge: catastrophic forgetting~\cite{mccloskey1989catastrophic,ratcliff1990connectionist}, existing methods can be divided into three categories: memory-based~\cite{riemer2019learning,buzzega2020dark,caccia2022new}, regularization-based~\cite{li2017learning,kirkpatrick2017overcoming,lin2022towards}, and dynamic architecture~\cite{douillard2022dytox,rusu2016progressive,wang2022learning,9477031}.
In this paper, we choose two classic settings in continual learning~\cite{9915459}: class incremental (class-il, models un-know task id) and task incremental (task-il, models know task id).
Memory-based continual learner shows superior performance among them in either class-il or task-il settings without expanding model size~\cite{buzzega2020dark,zhou2023deep}.
Hence, our research is focused on robust continual learners based on it.

What's more, some works~\cite{wu2021pretrained,yoon2019scalable,farquhar2018towards} focus on the robustness of the continual learning under varying experimental conditions, such as task-order, memory constraints, compute constraints or time constraints, and others focus on the robustness of continual learning models against backdoor attacks~\cite{umer2020targeted} or privacy
preservation~\cite{hassanpour2022differential}.
Several recent studies~\cite{khan2022adversarially,guoattacking,khan2022susceptibility} have identified the vulnerability of continual learning models to adversarial attacks, meanwhile applying adversarial sample techniques to stored data can mitigate catastrophic forgetting in continual learning, as observed in recent studies~\cite{wang2022improving,kumari2022retrospective}.
Chen~\etal~\cite{chen2022queried} first tries to enhance the adversarial robustness of continual learning models by combining LwF~\cite{li2017learning} with adversarial training and using lots of unlabeled data.
However, 
Differing from~\cite{chen2022queried}, we conduct an in-depth analysis of the main challenges in achieving adversarial robustness in continual learning and propose a solution that does not require additional data.

\noindent \textbf{Adversarial defense.}
Deep neural networks usually are vulnerable to adversarial examples~\cite{szegedy2013intriguing,duan2021advdrop,zhang2022towards}.
There are abundant adversarial defense methods to improve the models' adversarial robustness~\cite{akhtar2018threat,bai2021recent}.
Adversarial training has been proven the most effective way among them~\cite{zhang2019theoretically,croce2021robustbench,cui2021learnable,madry2018towards,maini2020adversarial,pang2022robustness,pang2021bag,9761760}, which leverages adversarial examples as training data.
Nowadays, adversarial robustness papers mainly focus on the ideal experimental setting, Wu~\etal~\cite{wu2021adversarial} take into account that data in the real world often have long-tailed distributions, and Shao~\etal~\cite{shao2020open} puts the problem of adversarial robustness in the open world.
Moreover, some studies~\cite{chou2022continual,rostami2021detection} show that continual algorithms can facilitate rapid model adaptation to new attack methods.
However, most of them are designed for single-task learning scenarios, and their effectiveness in continual learning scenarios remains largely unexplored.
Different from prior works, we delve into how to improve model's s adversarial robustness in class and task incremental settings.
\section{Baselines and Their Inherent Problems}
\label{sec:analysis}
\begin{table*}[t]
    \caption{A systematic experimental analysis of the combination of adversarial training (more specifically, Vanilla AT~\cite{madry2018towards}) and memory-based continual learning algorithms. See Sec. 3.2 for a detailed summary of observations. The formula definitions of FAA (final average accuracy), CRD (final average clean relative decrease), FRI (forgetting relative increase of clean data), and RRD (final average Robust Relative Decrease compared with the joint model), please see Sec.~\ref{sec:metrics}. Adversarial data are generated by PGD-20.
    For convenience, CRD, FRI, and RRD in the table are the averages of task incremental setting and class incremental setting.}.
    \label{Analysis}
    \centering 
    \resizebox{\textwidth}{!}{
    \begin{tabular}{cccrrrrrrrrlll}
    \toprule
    \multirow{3}{*}{\textbf{Buffer Size}} &
    \multirow{3}{*}{\textbf{Method}} &
    \multirow{3}{*}{\textbf{W/ AT}} &
    \multicolumn{4}{c}{\textbf{Class Incremental Setting}} &
    \multicolumn{4}{c}{\textbf{Task Incremental Setting}} &
    \multirow{3}{*}{\textbf{CRD$\downarrow$}} &
    \multirow{3}{*}{\textbf{FRI$\downarrow$}} &
    \multirow{3}{*}{\textbf{RRD$\downarrow$}} \\
    ~&~&~&\multicolumn{2}{c}{\textbf{Clean Data	}}&
    \multicolumn{2}{c}{\textbf{Adversarial Data}}&
    \multicolumn{2}{c}{\textbf{Clean Data}}&
    \multicolumn{2}{c}{\textbf{Adversarial Data}}&~&~&~\\
    ~&~&~&\textbf{FAA$\uparrow$ }&\textbf{Forgetting$\downarrow$}&
    \textbf{FAA$\uparrow$ }&\textbf{Forgetting$\downarrow$}&
    \textbf{FAA$\uparrow$ }&\textbf{Forgetting$\downarrow$}&
    \textbf{FAA$\uparrow$ }&\textbf{Forgetting$\downarrow$}&~&~&~\\
    \midrule
     \multirow{2}{*}{-}&\multirow{1}{*}{\textbf{Joint}} & $\usym{2717}$  & $91.81_{\pm{0.07}}$ & - & $0.00_{\pm{0.00}}$ & - & $98.17_{\pm{0.12}}$ & - & $0.00_{\pm{0.00}}$ &- &\multirow{2}{*}{$7.92$} &
     \multirow{2}{*}{-}&\multirow{2}{*}{-}\\
    \cmidrule(r){3-11}
     ~&(upper bound)& $\usym{2713}$ & $79.33_{\pm{0.36}}$ & - & $50.93_{\pm{1.28}}$ & - & $94.80_{\pm{0.30}}$ & - & $74.63_{\pm{1.13}}$ &- &~&~&~\\
    \midrule 
     \multirow{2}{*}{- }&\multirow{1}{*}{\textbf{SGD}}& $\usym{2717}$ & $19.66_{\pm{0.07}}$ & $96.39_{\pm{0.67}}$ & $5.17_{\pm{2.58}}$ & $33.31_{\pm{5.19}}$ & $66.84_{\pm{6.50}}$ & $37.42_{\pm{7.39}}$ & $4.44_{\pm{3.78}}$ &$33.34_{\pm{3.49}}$ &\multirow{2}{*}{$0.91$} &
     \multirow{2}{*}{$-3.73$}&\multirow{2}{*}{$41.10$}\\
     \cmidrule(r){3-11}
     ~& (lower bound) & $\usym{2713}$  & $19.34_{\pm{0.05}}$ &$91.93_{\pm{0.20}}$ & $15.65_{\pm{0.18}}$ & $69.71_{\pm{1.77}}$ & $65.33_{\pm{5.91}}$& $34.43_{\pm{7.16}}$ & $27.71_{\pm{1.88}}$ &$54.56_{\pm{3.47}}$ &~&~&~\\
    \midrule 
     \multirow{2}{*}{200}& \multirow{6}{*}{\textbf{ER}} &$\usym{2717}$  & $48.80_{\pm{1.84}}$ & $60.00_{\pm{2.25}}$ & $0.27_{\pm{0.18}}$ & $12.60_{\pm{2.13}}$ & $92.89_{\pm{0.26}}$ & $5.00_{\pm{0.26}}$ & $0.01_{\pm{0.01}}$ &$13.78_{\pm{0.41}}$ &\multirow{2}{*}{$14.51$} &
     \multirow{2}{*}{$12.91$}&\multirow{2}{*}{$31.70$}\\
    \cmidrule(r){3-11}
    ~&~& $\usym{2713}$  & $28.18_{\pm{0.69}}$ & $80.58_{\pm{1.05}}$ & $17.86_{\pm{0.29}}$ & $69.58_{\pm{1.15}}$ & $84.49_{\pm{0.61}}$ & $10.23_{\pm{0.98}}$ & $44.30_{\pm{1.05}}$ &$36.89_{\pm{1.11}}$ &~&~&~\\
    \cline{1-1}\cmidrule(r){3-14}
     \multirow{2}{*}{5120}&~& $\usym{2717}$ & $83.34_{\pm{1.38}}$ & $15.35_{\pm{1.71}}$ & $0.08_{\pm{0.11}}$ & $13.65_{\pm{2.34}}$ & $96.84_{\pm{0.17}}$ & $0.71_{\pm{0.19}}$ & $0.06_{\pm{0.04}}$ &$21.83_{\pm{8.69}}$ &\multirow{2}{*}{$13.53$} &
     \multirow{2}{*}{$12.11$}&\multirow{2}{*}{$18.80$}\\
    \cmidrule(r){3-11}
    ~& ~&$\usym{2713}$  & $61.88_{\pm{0.74}}$ & $37.72_{\pm{0.73}}$ & $27.28_{\pm{0.56}}$ & $41.66_{\pm{1.22}}$ & $91.24_{\pm{0.19}}$ & $2.56_{\pm{0.12}}$ & $56.59_{\pm{0.88}}$ &$19.34_{\pm{1.11}}$&~&~&~\\
    \midrule
     \multirow{2}{*}{200}&\multirow{6}{*}{\textbf{DER}} & $\usym{2717}$  & $62.71_{\pm{2.76}}$ & $38.75_{\pm{4.11}}$ & $0.01_{\pm{0.00}}$ & $7.42_{\pm{2.13}}$ & $91.32_{\pm{0.83}}$ & $6.95_{\pm{1.44}}$ & $0.09_{\pm{0.05}}$ &$7.39_{\pm{2.18}}$ &\multirow{2}{*}{$17.72$} &
     \multirow{2}{*}{$15.73$}&\multirow{2}{*}{$34.80$}\\
    \cmidrule(r){3-11}
    ~&~& $\usym{2713}$  & $35.79_{\pm{2.85}}$ & $65.67_{\pm{4.16}}$ & $16.44_{\pm{0.21}}$ & $52.84_{\pm{3.71}}$ &$82.79_{\pm{0.21}}$ & $11.48_{\pm{0.66}}$ & $39.52_{\pm{0.37}}$ &$40.64_{\pm{1.38}}$ &~&~&~\\
    \cline{1-1}\cmidrule(r){3-14}
    \multirow{2}{*}{5120}& ~& $\usym{2717}$&$83.99_{\pm{0.99}}$ & $10.57_{\pm{1.51}}$ & $0.07_{\pm{0.07}}$ & $7.05_{\pm{2.03}}$ & $95.67_{\pm{0.13}}$ & $1.63_{\pm{0.13}}$ & $0.05_{\pm{0.05}}$ &$7.39_{\pm{0.48}}$ &\multirow{2}{*}{$16.74$} &
     \multirow{2}{*}{$15.66$}&\multirow{2}{*}{$24.32$}\\
    \cmidrule(r){3-11}
    ~& ~&$\usym{2713}$ & $56.58_{\pm{3.87}}$ &$40.40_{\pm{4.07}}$& $23.26_{\pm{0.44}}$ & $45.06_{\pm{1.48}}$ & $89.62_{\pm{0.67}}$ & $3.12_{\pm{0.23}}$ & $53.66_{\pm{2.01}}$ & $22.95_{\pm{3.68}}$ &~&~&~\\
    \midrule
     \multirow{2}{*}{200}&\multirow{6}{*}{\textbf{DER++}} & $\usym{2717}$  & $65.55_{\pm{2.81}}$ & $32.17_{\pm{4.26}}$ & $0.24_{\pm{0.32}}$ & $7.38_{\pm{3.37}}$ & $92.11_{\pm{0.55}}$ & $8.85_{\pm{2.98}}$ & $0.22_{\pm{0.33}}$ &$10.45_{\pm{0.94}}$ &\multirow{2}{*}{$14.47$} &
     \multirow{2}{*}{$8.24$}&\multirow{2}{*}{$37.14$}\\
    \cmidrule(r){3-11}
    ~& ~&$\usym{2713}$  & $45.43_{\pm{3.28}}$ & $46.67_{\pm{5.09}}$ & $14.46_{\pm{1.35}}$ & $35.09_{\pm{2.70}}$ & $83.29_{\pm{1.07}}$ & $10.83_{\pm{1.77}}$ & $36.82_{\pm{0.59}}$ &$34.09_{\pm{2.23}}$&~&~&~\\
    \cline{1-1}\cmidrule(r){3-14}
     \multirow{2}{*}{5120}&~& $\usym{2717}$ & $85.74_{\pm{0.65}}$ & $7.32_{\pm{0.88}}$ & $1.51_{\pm{1.27}}$ & $6.10_{\pm{0.53}}$ & $96.13_{\pm{0.53}}$ & $1.14_{\pm{0.35}}$ & $0.82_{\pm{0.52}}$ &$6.06_{\pm{0.49}}$ &\multirow{2}{*}{$10.63$} &
     \multirow{2}{*}{$8.21$}&\multirow{2}{*}{$22.18$}\\
    \cmidrule(r){3-11}
    ~& ~&$\usym{2713}$  & $68.67_{\pm{1.11}}$ & $23.68_{\pm{2.66}}$ & $26.95_{\pm{0.29}}$ & $27.68_{\pm{2.48}}$ & $91.95_{\pm{0.33}}$ & $1.20_{\pm{0.90}}$ & $54.26_{\pm{0.41}}$ &$16.87_{\pm{1.72}}$&~&~&~\\
    \midrule
     \multirow{2}{*}{200}&\multirow{6}{*}{\textbf{X-DER}} & $\usym{2717}$  & $60.96_{\pm{1.09}}$& $12.29_{\pm{0.09}}$& $0.00_{\pm{0.00}}$& $3.84_{\pm{1.18}}$& $94.33_{\pm{0.63}}$& $2.49_{\pm{0.17}}$& $0.13_{\pm{0.03}}$& $2.59_{\pm{0.64}}$ &\multirow{2}{*}{$20.23$} &
     \multirow{2}{*}{$7.65$}&\multirow{2}{*}{$24.02$}\\
    \cmidrule(r){3-11}
    ~&~& $\usym{2713}$  & $34.04_{\pm{0.90}}$& $25.13_{\pm{4.03}}$& $16.82_{\pm{0.98}}$& $27.84_{\pm{4.16}}$& $80.8_{\pm{0.66}}$& $4.96_{\pm{1.66}}$& $60.83_{\pm{0.56}}$& $10.99_{\pm{0.81}}$ &~&~&~\\
    \cline{1-1}\cmidrule(r){3-14}
    \multirow{2}{*}{5120}&~& $\usym{2717}$ & $69.93_{\pm{0.63}}$& $4.33_{\pm{0.26}}$& $0.21_{\pm{0.12}}$& $4.52_{\pm{1.39}}$& $96.94_{\pm{0.02}}$& $0.29_{\pm{0.04}}$& $0.13_{\pm{0.02}}$& $2.04_{\pm{0.43}}$ &\multirow{2}{*}{$24.96$} &
     \multirow{2}{*}{$18.52$}&\multirow{2}{*}{$21.27$}\\
    \cmidrule(r){3-11}
    ~&~&$\usym{2713}$& $34.29_{\pm{0.65}}$& $38.92_{\pm{0.03}}$& $19.24_{\pm{0.10}}$& $38.79_{\pm{0.41}}$& $82.67_{\pm{1.68}}$& $2.74_{\pm{0.83}}$& $64.11_{\pm{0.12}}$& $8.14_{\pm{0.98}}$ &~&~&~\\
\bottomrule
\end{tabular}}
\end{table*}
\subsection{Preliminaries}

\noindent \textbf{Memory-based continual learning.}
Following~\cite{buzzega2020dark,boschini2022class},
we consider the scenario of continual learning across $T$ distinct classification tasks.
The data distribution for the $t$-th task ($1 \le t \le T$) is $D_t$, $x_t$ is the input sample of $t$-th task, and $y_t$ is the corresponding label.
A model $\mathrm{M}_{\theta}(\cdot)$ with parameters $\theta$ is optimized sequentially in the given task order.
We denote the output logits as $h_{\theta}(x_t) \in \mathbb{R}^{C}$, where $h_{\theta}(x_t)=\mathrm{M}_{\theta}(x_t)$, and $C$ is the total number of categories.
And $f_{\theta}(x_t)\triangleq \mathrm{softmax}(h_{\theta}(x_t))$.

In this paper, we share a classification header during the total training phase.
We divide the classifier head into three parts according to the task order: past classes $p$, current classes $c$, and unseen classes $u$.
We set $h_{\theta}(x_t) = [h_{\theta}(x_t)_{1}, ..., h_{\theta}(x_t)_{i}, ..., h_{\theta}(x_t)_{C}]$, where $h_{\theta}(x_t)_{i}$ means the logit corresponding to the $i$-th class of sample $x_t$.
\begin{equation}
\begin{aligned}
p =& \{1,2, ..., (t-1)*b\},\\ 
c =& \{(t-1)*b+1, (t-1)*b+2, ... ,t*b\},\\
u =& \{t*b+1,t*b+2,...,C\},
\end{aligned}
\end{equation}
where $b$ is the number of classes of each task.
So we can divide $h_{\theta}(x_t)$ into three parts: $h_{\theta}(x_t)_{p}$, $h_{\theta}(x_t)_{c}$, and $h_{\theta}(x_t)_{u}$.
$h_{\theta}(x_t)_{p}$ means the logits of past classes of $x_t$, $h_{\theta}(x_t)_{c}$ means the logits of current classes of $x_t$, and $h_{\theta}(x_t)_{u}$ means the logits of future classes of $x$.

Continual learner aims to correctly classify data of all observed tasks, and its objective can be formalized as:
\begin{equation}
    \begin{aligned}{
\mathop{\arg\min}_{\theta} \sum_{t=1}^{T}  \mathcal{L}_{t}, \ \ \mathrm{where} \ \  \mathcal{L}_{t} \triangleq \mathbb{E}_{(x_t,y_t)\sim D_t}[ \ell (f_{\theta}(x_t),y_t)],}
    \end{aligned}
\end{equation}
where $\ell$ is the task-specific loss.
The unavailability of past task data poses great challenges for continual learning.
However, direct sequential optimization can lead to severe catastrophic forgetting.
To address the above challenge, memory-based continual learners usually store a few past task data and replay them during the current task training phase~\cite{buzzega2020dark}, and become leading-edge approaches.
When training the $t$-th task, its loss function can be formalized as the following:
\begin{equation}
\begin{aligned}
\mathcal{L}_{t} \triangleq \mathbb{E}_{(x,y)\sim D_{t} \cup \mathcal{M}} [\ell (f_{\theta}(x),y)], 
\end{aligned}
\end{equation}
where $\mathcal{M}$ is the episodic memory of the current task, which preserves a small subset of observed past data $\langle x_{\mathcal{M}},y_{\mathcal{M}} \rangle$.

\noindent \textbf{Adversarial training.}
Adversarial training is currently the most reliable way~\cite{bai2021recent} to improve the adversarial robustness of a model.
Following~\cite{madry2018towards}, it can be formally defined as a min-max optimization problem:
\begin{equation}
\mathop{\arg\min}_{\theta} \mathbb{E}_{(x, y) \sim D}\left(\max_{\delta \in S} { \ell}(f_{\theta}(x+\delta), y)\right),
\label{eq:at}
\end{equation}
where $\delta$ is an adversarial perturbation, $D$ is data distribution, and $S$ is a perturbation set.
$S =\left \{ \delta | \left \| \delta \right \|_{p} \le \epsilon  \right \}$, $\epsilon$ is the perturbation size. The goal of adversarial training is to correctly classify all adversarial examples ($\widetilde{x} = x+\delta$).

Here we integrate memory-based continual learner and adversarial training as adversarial robust memory-based continual learner baselines, and the new optimization objective during the training stage of $t$-th task is:
\begin{equation}
\mathop{\arg\min}_{\theta} \mathbb{E}_{(x, y) \sim D_{t} \cup \mathcal{M}}\left(\max_{\delta \in S} { \ell}(f_{\theta}(x+\delta), y)\right).
\label{eq:total}
\end{equation}

\subsection{Vanilla Solutions}
To reveal the challenges inherent in adversarial robust continual learning, we present a detailed and systematic analysis starting from popular memory-based continual learning methods: ER~\cite{riemer2019learning}, DER~\cite{buzzega2020dark}, DER++~\cite{buzzega2020dark} and X-DER~\cite{boschini2022class} with different memory buffer sizes on the Split-CIFAR10~\cite{krizhevsky2009learning} dataset.
Both class incremental and task incremental settings are considered.
Given the computational cost, we employ the classic and effective Vanilla AT~\cite{madry2018towards} (termed as ``AT'' in the following) algorithm in the adversarial training section of our work.
For comparison, we present an upper bound given by training all tasks simultaneously (termed as ``Joint'') and a lower bound given by training all tasks sequentially with SGD without any anti-forgetting strategy (termed as ``SGD'').
Following~\cite{buzzega2020dark}, final average accuracy (FAA) and forgetting in the scenarios of both clean and adversarial data are used as metrics.
Additional metrics include the clean relative decrease (CRD), forgetting relative increase (FRI), and robust relative decrease versus joint model (RRD).
CRD assesses the relative drop in clean FAA of a continual learner after adversarial training.
Similarly, FRI evaluates the mean relative increase of clean-forgetting under different settings after adversarial training.
RRD measures the mean gap between joint model and continual models in relative increase of adversarial robustness after adversarial training.
For more baseline and metric details, see Sec.~\ref{sec:metrics}.

Results in Table~\ref{Analysis} validate the observations in Sec.~\ref{sec:intro}.
When integrated with adversarial training, continual learning models improve robustness while reducing the clean performance (CRDs are all positive); and adversarial training accelerates the forgetting of clean samples (FRIs of memory-based continual learnings are all positive), while robustness also forgets.
Furthermore, the results show that a large memory size leads to an overall improvement in performance both on FAA/forgetting for clean and adversarial data.

In summary, the results indicate that adversarial training is not suitable for direct integration with continual learning models.
Initially, we notice that for the same amount of data and super-parameters, the improvement in robustness of the continual learning model with adversarial training is quite inferior to that of the joint model (RRDs are all positive).
Meanwhile, when DER/DER++ and X-DER are combined with adversarial training, they show favorable stability in the stability-plasticity dilemma, disproportionately inhibiting learning ability on the new task.
In contrast, the combination of ER and adversarial training shows strong robustness performance, especially in the class incremental setting with a $200$ buffer size.
Owing to its straightforward yet efficient results, we select ER as the foundational continual learning model to assess the efficacy of various adversarial training methodologies in continual scenarios, as detailed in Sec~\ref{sec:experiment}.
For hyper-parameter specifics of the baseline models in Table~\ref{Analysis}, please see the Appendix.

\subsection{The Inherent Problems}
We conduct a deeper analysis of the aforementioned phenomena to enhance the robust continual learning models.
In the process, we identified two significant challenges associated with the direct integration of adversarial training and continual learning: accelerated forgetting for continual learning and gradient obfuscation for adversarial training.

\noindent \textbf{Negative impact of adversarial training on continual learning: accelerated forgetting.}
As mentioned above, adversarial training accelerates the forgetting of continual learners of clean data.
Specifically, adversarial training with only stored past samples has the effect of alleviating catastrophic forgetting of clean data, but adversarial training with all samples accelerates forgetting.
This is due to the gradient amplification property of the adversarial sample.
According to~\cite{madry2018towards}, 
\begin{equation}
\begin{aligned}
\widetilde{x} = x+\delta, \ \ {\rm where} \ \delta \triangleq \arg\max_{\delta \in S} { \ell}(f_{\theta}(x+\delta), y),
\end{aligned}
\end{equation}
adversarial sample raises the logits corresponding to the other categories, and data from current task $t$ are vulnerable data prone to attack as past categories, especially in the early training stages.
We set $\widetilde{x_{t}}$ is the adversarial example generated based on clean training sample $x_{t}$ of task t.
It can be naturally implied that the adversarial sample $\widetilde{x_{t}}$ of the current task has a higher logit in the other categories compared to the clean sample $x_{t}$, while its own true category corresponds to a lower logit.
Namely,
\begin{equation}
\begin{aligned}
h_{\theta}(\widetilde{x_{t}})_p > h_{\theta}(x_{t})_p, \ h_{\theta}(\widetilde{x_{t}})_{c} < h_{\theta}(x_{t})_{c}.
\end{aligned}
\end{equation}
We choose the cross-entropy as the loss function, and the negative gradients of the current data $x_t$ to the past task are increased. Namely,
\begin{equation}
\label{eq:gra0}
\left[\frac{\partial \ell}{\partial h_{\theta}(\widetilde{x_{t}})} \right]_{p} = f_{\theta}(\widetilde{x}_{t})_{p} >  f_{\theta}({x}_{t})_{p} = \left[\frac{\partial \ell}{\partial h_{\theta}({x_{t}})} \right]_{p}, 
\end{equation}
which means adversarial examples of current data result in greater negative gradients towards the past tasks, especially in the early training stages of the new task.
Similarly, we can speculate that adversarial examples of stored past data lead to a greater negative gradient toward current tasks, thus alleviating catastrophic forgetting.
However, the number of stored samples is limited; thus, using the entire stored and current data for adversarial training will still accelerate forgetting.

\begin{figure}
\label{fig:gradient_norm}
\centering





    


\includegraphics[width=1\linewidth]{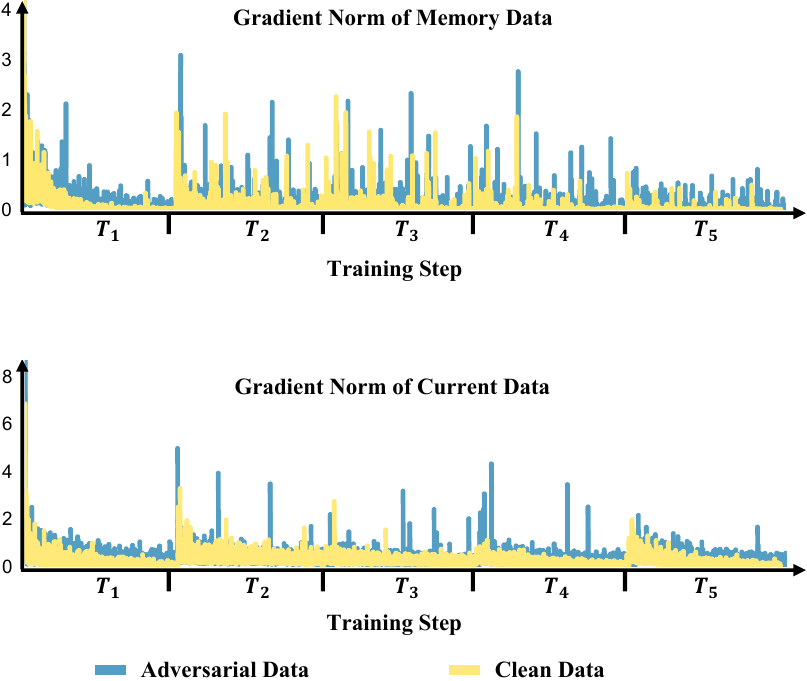}
\caption{Gradient norms of adversarial data and clean data during the training stage. We experiment ER+AT with a $200$ buffer size on the Split-CIFAR10 dataset.} 
\label{fig:gn}
\end{figure}
We perform statistics on the gradient norm brought to the classification head in the training process of ER+AT for the adversarial and clean samples in the dataset Split-CIFAR10, with a buffer size of 200.
As shown in Fig.~\ref{fig:gn}, both current task data and memory data show that adversarial data bring greater gradients than clean data.

\begin{figure}[t]
  \centering
  \includegraphics[width=0.8\linewidth]{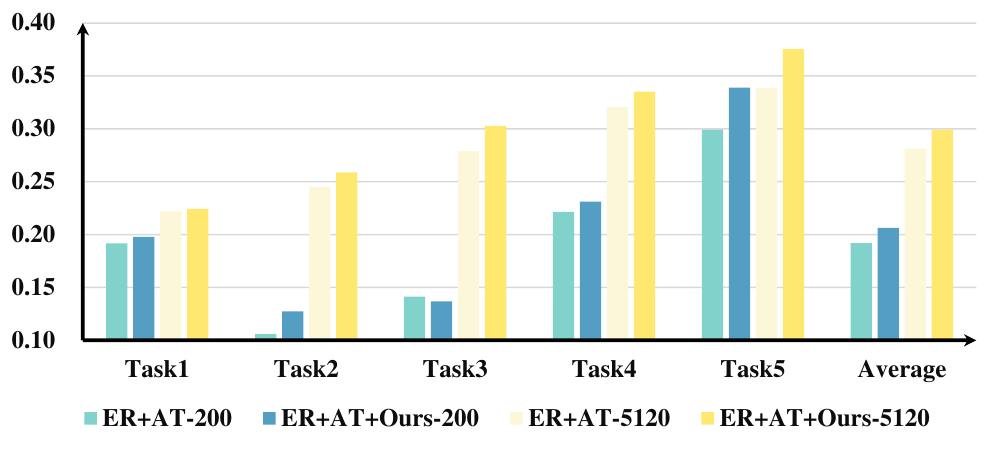}
  \caption{The gradient cosine similarities between robust joint model and robust continual model, which indicates a special gradient obfuscation only in continual learning. We experiment with buffer size=200/5,120 on the Split-CIFAR10 dataset. Gradient obfuscation weakens as the buffer size gets larger, and our method also mitigates gradient obfuscation.}
  \label{fig:gradient_similarity}
\end{figure}
\begin{figure*}[t]
  \centering
  \includegraphics[width=0.99\linewidth]{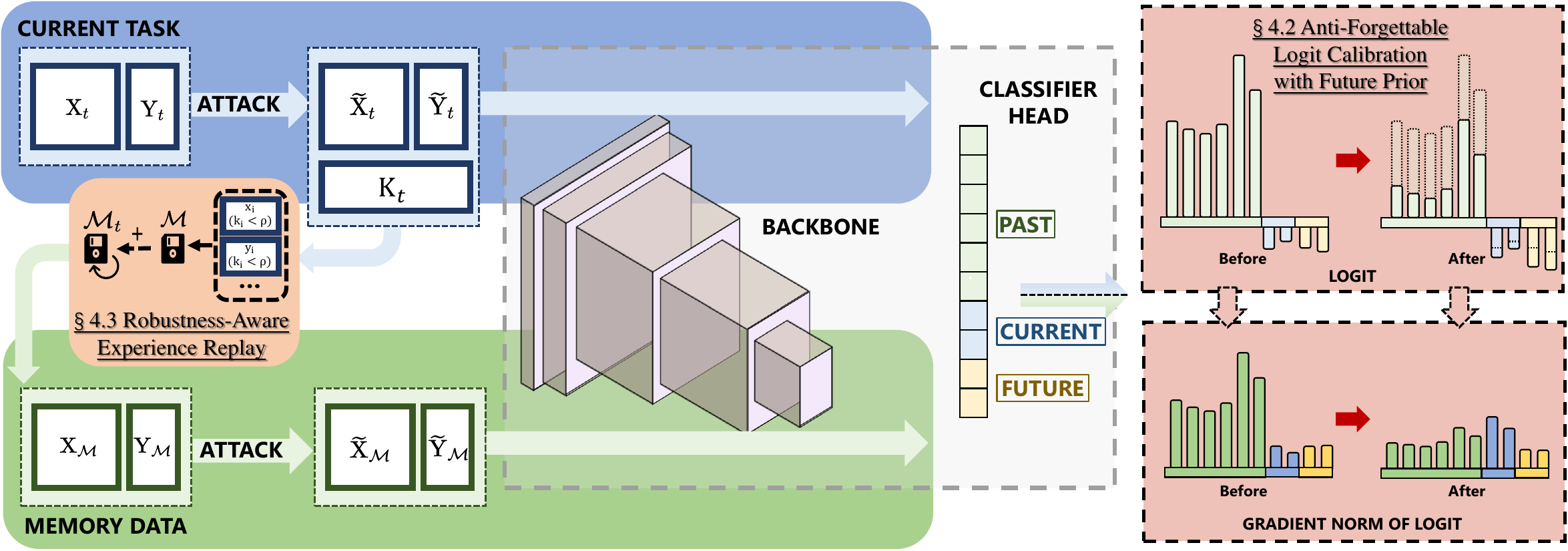}
  \caption{The pipeline of the adversarial robust memory-based continual learner.
  We first generate the adversarial samples $\widetilde{X}_{t}$, $\widetilde{X}_{\mathcal{M}}$ and robust difficulty coefficients $K_{t}$ for the current task $t$ data ($X_{t}, Y_{t}$) and past data ($X_{\mathcal{M}}, Y_{\mathcal{M}}$) in memory $\mathcal{M}$. Then all of them are fed into the continual learning model for training.
  To mitigate the forgetting acceleration due to adversarial examples, we use the logit adjustment strategy (Sec. 4.2) to reduce the negative gradient of the new task data on the parameters of the old task by adjusting the sample logit layer, especially during task switching.
  In addition, to alleviate gradient obfuscation, we filter the current data according to the robust difficulty factor $K_{t}$ when saving the data into memory $\mathcal{M}$ (Sec. 4.3).
  }
  \label{Pipeline}
\end{figure*}
\noindent \textbf{Negative impact of continual learning on adversarial training: gradient obfuscation.}
Gradient obfuscation~\cite{athalye2018obfuscated} refers to the deliberate or accidental obscuring or confusing of gradient information caused by non-differentiable operators, unstable values, randomness in the testing phase, or deep computing.
Unlike the previous instances, we observe a new particular ``shattered gradients~\cite{athalye2018obfuscated}'' of gradient obfuscation in adversarial training only under continual settings, which results in inferior robustness improvements compared to the joint model.

Gradient obfuscation under continual learning specifically means gradient information is biased due to limited replay data.
When constructing class boundaries for the old and new tasks, the model has only a small number of old task samples, which results in overfitting to the data in memory and direction-biased gradients when generating adversarial examples.
As shown in Fig.~\ref{fig:gradient_similarity}, we calculate the gradient directions of the adversarial robust joint model and continual model, and we find that there is a significant difference between the gradient directions of the continual learner and the joint learner, which leads to lower adversarial robustness for continual learners.


\section{Method}
\label{sec:method}
\subsection{Overview}

To mitigate the forgetting acceleration caused by adversarial samples, we propose anti-forgettable logit calibration (AFLC);
Additionally, to reduce gradient obfuscation due to limited past data, we present a new robustness-aware data selection for experimental replay (RAER).
The specific process is shown in Fig.~\ref{Pipeline}. When the model learns a new task, data logits are adjusted according to the task order (past, current, or future) to mitigate forgetting.
Subsequently, the data from the current task will be filtered into memory according to the robustness difficulty factor $k$ and replayed when the next task is learned.

\subsection{Anti-Forgettable Logit Calibration}
We propose a simple but effective logit calibration strategy for robust adversarial training.
It mitigates the accelerated forgetting caused by adversarial samples through subtracting larger calibration values for $h_{\theta}(\widetilde{x})_{p}$ than $h_{\theta}(\widetilde{x})_{c}$, where $h_{\theta}(\widetilde{x})$ can be $h_{\theta}(\widetilde{x_t}))$ or $h_{\theta}(\widetilde{x}_{\mathcal{M}})$.
Namely,
\begin{equation}
\begin{aligned}
\label{eq:logit-adjust}
h^{\mathrm{lc}}_{\theta}(\widetilde{x})_{i} = h_{\theta}(\widetilde{x})_{i} - \mathrm{v}_i,
\end{aligned}
\end{equation}
where $\mathrm{v}_i$ means the task-order-based logit calibration value added to class $i$, $\mathrm{v} \geq 0$ and $\mathrm{v}_{p} > \mathrm{v}_{c}$.
In this way, for stored past data, we can maintain the advantage of adversarial samples to mitigate catastrophic forgetting; For current task data, we can attenuate their negative gradients to past classes, thus mitigating forgetting acceleration.

When training $t$-th task, the ultimate optimization goal is:
\begin{equation}
\label{eq:total2}
    \mathop{\arg\min}_{\theta} \mathbb{E}_{(x, y) \in D_{t} \cup M\left(\max_{\delta \in S} { \ell}(f^{\mathrm{lc}}_{\theta}(x+\delta), y)\right), 
}\end{equation}
\begin{equation}
\label{eq:total3}
f^{\mathrm{lc}}_{\theta}(x+\delta)_i = \frac{e^{h_{\theta}(\widetilde{x})_{i} -  \mathrm{v}_i}}{\sum_{j=1}^C e^{h_{\theta}(\widetilde{x})_{j} - \mathrm{v}_j}}, \ \ \ \mathrm{for}\ i=1,2,\dots,C.
\end{equation}
Due to $\mathrm{v}_{p} > \mathrm{v}_{c}$, the negative gradients of the current data $x_t$ to the past task are reduced. Namely,
\begin{equation}
\label{eq:gra1}
\left[\frac{\partial \ell}{\partial h^{\mathrm{lc}}_{\theta}(\widetilde{x})} \right]_{p} = f^{\mathrm{lc}}_{\theta}(\widetilde{x}_{t})_{p} <  f_{\theta}(\widetilde{x}_{t})_{p},
\end{equation}
and the negative gradients of the past data to the current task increase. Namely,
\begin{equation}
\label{eq:gra2}
\left[\frac{\partial \ell}{\partial h^{\mathrm{lc}}_{\theta}(\widetilde{x_{\mathcal{M}}})} \right]_{c} = f^{\mathrm{lc}}_{\theta}(\widetilde{x}_{\mathcal{M}})_{c} >  f_{\theta}(\widetilde{x}_{\mathcal{M}})_{c},
\end{equation}
which can effectively mitigate forgetting due to adversarial examples.

For the specific implementation, $\mathrm{v}_i = -\log(\frac{n_i}{\sum_{j=1}^C n_j }) - \alpha_i$, where $\alpha_i$ is a hyper-parameter, and $n_i$ is total number of samples of class $i$ in memory and current data.
Note that the model does not do anything with logit during inference time.

\textbf{Further Prior ($FP$) adjustment.}
When all tasks share the classification head, we realize that making logit changes to past and current classes only is equivalent to negative gradient increases for future classes.
So we make a further prior adjustment: for future classes,
$ \mathrm{v}_i = \frac{1}{t*b}\sum^{t*b}_{j=1}(\mathrm{v}_{j})$, where $t$ means the current task order and $b$ is the number of classes of each task.

\textbf{Comparisions with logit adjustment.}
Logit adjustment~\cite{menon2021longtail,zhao2022adaptive} is often used in long-tail learning to reduce the negative influence of tail classes from head classes.
Unlike them, we consider future tasks which are specific to continual learning.

X-DER~\cite{boschini2022class} also adjusts the negative gradients of the new task data affecting the past tasks
by updating the new task with the logit masking to avoid the negative impact on the old task classifier parameters.
Logit masking can be seen as a special case of our method at the softmax layer, which can be viewed as $\mathrm{v}_{p} = \mathrm{v}_{c} = 0$ for past data, and $\mathrm{v}_{p} = +\infty$, $ \mathrm{v}_{c} = 0$ for current data.
However, their extreme margin v inhibits the model's ability to learn new knowledge.
X-DER further adds a safeguard threshold to limit the activations of past and future heads, which also can be combined with us.
The experimental section shows that replacing logit masking with AFLC in X-DER can achieve better performance.


\subsection{Robustness-Aware Experience Replay}
By leveraging attack difficulty, we store data that own adversarial robustness and maintain the diversity of past data to mitigate gradient obfuscation, instead of storing data near or across their decision boundary~\cite{chaudhry2018riemannian}.

Specifically, we design a data selection strategy based on the robustness difficulty factor $k$.
$k$ is the number of successful attacks during an adversarial example generation by iterative gradient-based attack methods (such as PGD~\cite{madry2018towards}). It can represent the attack difficulty and the distance of a sample to the task decision boundary.
A high value of $k$ indicates that the sample is vulnerable and close to decision boundaries, even overlapping with distribution areas of other classes.
While a small $k$ value implies this sample is robust, and far from the classification boundary.

\textbf{Memory-update process.}
During training, we use the PGD-$10$ algorithm to generate adversarial examples.
To determine the robustness difficulty factors for the batch data, we introduce difficulty factors of batch data as $K = \{k_{x_1}, k_{x_2}, ..., k_{x_i}, ..., k_{x_n}\}$,
where $k_{x_i}$ means the robustness difficulty factor of sample $x_i$, $ \ k_{x_i} \in [0,10], \  k_{x_i} \in \mathbb{N}$, and $n$ is the batch size.
We set a threshold $\rho$, only randomly select current task samples that satisfy $k < \rho$ into memory, and filter out specific difficult samples that overfit the current classification boundary as outliers by $k$.

Selecting data satisfying diversity and safety is more conducive to robust continual learning.
These samples can better represent the data distribution and are easier for the model to classify and remember correctly.
This contrasts with the practices commonly adopted in continual learning~\cite{chaudhry2018riemannian} and adversarial training~\cite{zhang2021_GAIRAT}, where samples near the decision boundary are often greater emphasis.

\noindent\textbf{Comparisons with other data selection-based continual learning methods.}
There are also some data selection-based methods in continual learning to mitigate catastrophic forgetting.
For example, GSS~\cite{aljundi2019gradient} maximizes data diversity using parameter gradients; ASER~\cite{shim2021online} uses Shapley value to select samples that are beneficial for latent decision boundary learning for both old and new classes; and more.
However, they all focus on continual learning settings with only clean samples.

Unlike previous work, RAER is designed to alleviate the gradient obfuscation in the continual robust setting.
\begin{figure*}[t]
  \centering
  \includegraphics[width=1.0\linewidth]{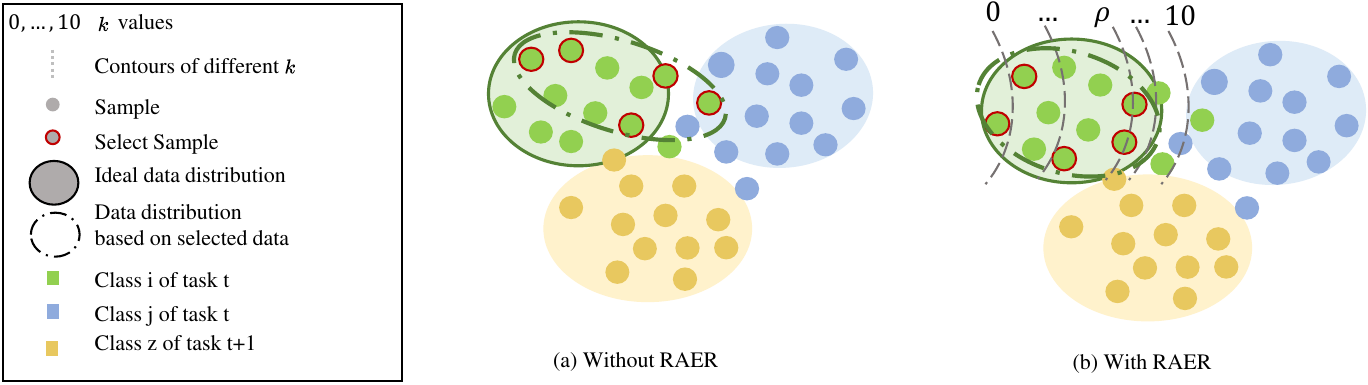}
  \caption{Schematic diagram of the RAER. A larger $k$ means that the sample is more vulnerable, so the closer it is to its task decision boundary. RAER can exclude vulnerable samples that over-fit the boundary of the current task, thus selecting samples that are more robustly safe and more representative of the data distribution.}
  \label{fig:raer}
\end{figure*}
As shown in \figurename~\ref{fig:raer}, RAER is effective because samples that are adversarial safe are selected and the model is more likely to remember them; data with too large k values are excluded, they are usually close to or even beyond the task boundary to which they belong, and selection among the remaining data can better depict the data distribution and alleviate gradient obfuscation.

In Sec.~\ref{sec:experiment}, we have also shown that existing continual learning data selection strategies combined directly with adversarial training do not achieve better performance than the ER algorithm (random selection), especially in terms of adversarial robustness, whereas our algorithm can effectively mitigate gradient obfuscation.
\begin{algorithm}
\caption{Memory-based Adversarial Robust Continual Learner}
\label{algorithm}
\KwIn{Model $\mathrm{M}$ with parameters $\theta$, memory $\mathcal{M}$, batch data $(X_1,Y_1), ..., (X_T,Y_T)$ respectively from different task distributions $\{\mathcal{D}_1, ..., \mathcal{D}_T\}$, step size of adversarial perturbations $\epsilon$, the number of task $t$ epochs $epoch_t$}
\KwResult{Final model $\mathrm{M}_{\theta_T}$}
$\mathcal{M} \leftarrow \{\}$\;
Random initialize $\theta_0$\;
    \For {$t = 1, ..., T$}
        {
        $\theta_t \leftarrow \theta_{t-1}$\;
        \For{$m=1, ..., epoch_t$}{
            Sampling a random batch $(X_{t},Y_{t})\sim \mathcal{D}_t$\;
            \If{$\mathcal{M} \neq \{\}$}{
                Sampling a random batch $(X_{\mathcal{M}},Y_{\mathcal{M}})\sim \mathcal{M}$\;
                $X_t \leftarrow [X_{t}, X_{\mathcal{M}}]$,
                $Y_t \leftarrow [Y_{t}, Y_{\mathcal{M}}]$\;}
            $\widetilde{X}_t, K_{t} \leftarrow$ PGD$(\theta_t,X_{t},Y_{t},\epsilon)$\;
            $h_{\theta_t}(\widetilde{X}_t) \leftarrow \mathrm{M}_{\theta_t}(\widetilde{X}_t)$\;
            $h^\mathrm{lc}_{\theta_t}(\widetilde{X}_t) \leftarrow$ AFLC$(h_{\theta_t}(\widetilde{X}_t))$ Eq.~(\ref{eq:logit-adjust})\;
            $\theta_{t}^{m+1} \leftarrow$ Update $(\theta_{t}^{m},h^\mathrm{lc}_{\theta_t}(\widetilde{X}_t),Y_{T})$ Eq.~(\ref{eq:total2})\;
            $\mathcal{M}_{t} \leftarrow $ RAER$(\mathcal{M},X_{t},K_{t})$\;
        }
        $\mathcal{M} \leftarrow \mathcal{M} \cup \mathcal{M}_{t}$\;
    }
\end{algorithm}

\subsection{Overall Objective}
To sum up, we design AFLC to reduce negative gradients of the current adversarial sample to past tasks, and RAER to select data that alleviate gradient obfuscation to store. We set $\alpha$ of AFLC is $3.5$ in practice and and $x_{\mathcal{M}}$ and $y_{\mathcal{M}}$ are selected by RAER with $\rho=5$.
When ER+AT combines with ours, the loss of task t is:
 \begin{equation}
 \begin{aligned}
 \label{eq:er+at+ours}
\mathcal{L}_{t} &\triangleq \  \mathrm{CE}(f^{lc}_{\theta}({\widetilde{x_t}}),y_{t})
 +\mathrm{CE}(f^{lc}_{\theta}({\widetilde{x}_{\mathcal{M}}}),y_{\mathcal{M}}).
 \end{aligned}
 \end{equation}
When ER+TRADES combines with ours, the loss of task t is:
 \begin{equation}
 \begin{aligned}
 \label{eq:er+trades+ours}
 \mathcal{L}_{t} &\triangleq \  \mathrm{CE}(f^{lc}_{\theta}({{x_t}}),y_{t})
 + \beta* \mathrm{KL}(f^{lc}_{\theta}({x_t}),f^{lc}_{\theta}({\widetilde{x_t})})\\
 &+\mathrm{CE}(f^{lc}_{\theta}({{x_{\mathcal{M}}}}),y_{\mathcal{M}})
 + \beta* \mathrm{KL}(f^{lc}_{\theta}({x_\mathcal{M}}),f^{lc}_{\theta}({\widetilde{x}_\mathcal{M})}),
 \end{aligned}
 \end{equation}
where $\beta$ of TRADES is $6.0$.

\section{Experiments}
\label{sec:experiment}

\subsection{Experimental Settings}
\noindent\textbf{Datasets.}
Following common adversarial training and continual learning works~\cite{pang2022robustness,boschini2022class,zhu2023improving}, we conduct systematic analytical experiments on the Split-CIFAR10~\cite{krizhevsky2009learning} dataset and validate our improvements on the Split-CIFAR10, Split-CIFAR100~\cite{krizhevsky2009learning}, and Split-Tiny-ImageNet~\cite{stanford2015tinyimagenet} datasets.
The Split-CIFAR10 contains ten classes, with $5,000$ training samples and $1,000$ test samples per class. 
Split-CIFAR100 consists of $100$ classes, each with a  set of $500$ training samples and a test set of $100$ samples.
In the continual learning setting, Split-CIFAR10 is divided into five binary classification tasks, and Split-CIFAR100 is divided into ten tasks,  each consisting of a ten-way classification task.
The Split-Tiny-ImageNet has $200$ classes, with $500$ samples per class for training and $50$ samples for validation and testing, respectively, and is split into ten tasks, where each task is a $20$-way classification task.

\noindent\textbf{Training details.}
Following common adversarial training settings, we set perturbations range of $8/255$ and step size of $2/255$ while generating adversarial samples.
In the training phase, following DER and X-DER, the learning rate is $0.1$, and the model architecture is ResNet18.
For Split-CIFAR10, Split-CIFAR100, and Split-Tiny-ImageNet, we perform random cropping with padding of four pixels and horizontal flipping for both the stream and buffer examples.
Here, we only consider how to achieve adversarial robust continual learners without large amounts of unlabeled data\footnote{The main external source for Chen~\etal~\cite{chen2022queried} is an 80M-TinyImage dataset, which has been withdrawn due to privacy violations.}.
We train all networks using the SGD optimizer, which is also consistent with DER.
The results of all experiments are run three times on different random seeds in Split-CIFAR10 and Split-CIFAR100 datasets and two times in Split-Tiny-ImageNet, and the mean and standard deviation are calculated.
Due to the size of the table, we only show the mean results in the paper and put the results with standard deviations in the Appendix.

\noindent\textbf{Baselines.}
\label{sec:baselines}
Due to the particularity of our task, our baselines comprise continual learning methods and adversarial training methods, e.g. ``ER+AT''.
For the part of continual learning baselines, we choose four popular continual learning algorithms, ER~\cite{riemer2019learning}, DER~\cite{buzzega2020dark}, DER++~\cite{buzzega2020dark}, and X-DER~\cite{boschini2022class} in the analysis section.
Furthermore, we combine our approach with two data selection-based continual learning methods: GSS~\cite{aljundi2019gradient} and ASER~\cite{shim2021online}, and a logit masking-based method X-DER to show our performance in the main results section.
ER randomly stores samples of past tasks and replays them in new tasks, achieving superior results without other operations;
DER and DER++ store logits of old data based on ER, further alleviating catastrophic forgetting by distilling knowledge from past tasks, and DER++ additionally utilizes labels of past data to be resistant to forgetting;
and X-DER embraces memory update and future preparation and uses logit masking, a special case of our AFLC, to reduce overweighting negative gradients of current data for past data.
GSS selects diverse samples based on gradients.
While ASER, also based on ER, utilizes the Shapley value to identify the most helpful data for mitigating forgetting.

For the part of adversarial robustness baselines, we choose four popular adversarial training algorithms: Vanilla AT~\cite{madry2018towards} (abbreviated as AT in our experiments), TRADES~\cite{zhang2019theoretically}, FAT~\cite{zhang2020attacks}, LBGAT~\cite{cui2021learnable}, and SCORE~\cite{pang2022robustness}.
AT adds the adversarial sample directly as training data, while TRADES adds a regular term that requires the adversarial sample to be consistent with the corresponding clean sample in logit outputs, both of which are currently strong robust baselines~\cite{pang2021bag}.
FAT chooses the adversarial sample that just succeeds in each attack to reduce clean accuracy decline in adversarial training.
LBGAT achieves both robustness and clean accuracy improvements by distilling the logit of the standard training model. 
SCORE employs local equivariance to describe the ideal robust model's behavior to achieve top-rank performance in both robust and clean data.

Given the expensive computation of exhaustively exploring permutations of various continual learning and adversarial training algorithms, we adopt ER as the foundational baseline in combination with adversarial training algorithms based on the simplicity and effectiveness of ER+AT~(Table~\ref{Analysis}), and choose AT and TRADES as adversarial training baselines in evaluate the effectiveness of our approach because of the superior performance of ER+AT and ER+TRADES in adversarial FAA (Table~\ref{MainCIFAR}). 

Both continual learning and adversarial training are hyper-parameter-sensitive domains. To reduce the workload of tuning parameters, we keep the hyper-parameters of the continual learning algorithm consistent with the DER code, and we keep the hyper-parameters of the adversarial training algorithm consistent with their original papers.

\textbf{Combined with different continual learning methods.}
When the continual algorithms are combined with Vanilla AT (AT), the input of its loss function only changes from clean samples to adversarial samples, so it will not be explained in detail.
\begin{itemize}{
 \item \textbf{ER+AT.}
 We set the learning rate as $0.1$, batch size as $32$, and the number of epochs per task as $50$.
 \item \textbf{GSS+AT.}
 We set the learning rate as $0.03$, batch size as $32$, and the number of epochs per task as $50$.
 \item \textbf{ASER+AT.}
 We set the learning rate as $0.1$, batch size as $32$, the maximum number of samples per class for random sampling as $1.5$, the number of nearest neighbors to perform ASER as $3$, and the number of epochs per task as $20$.
 \item \textbf{X-DER+AT.}
 We set the learning rate as $0.03$, batch size as $32$, m as $0.7$, alpha is $0.05$, beta is $0.01$, gamma as $0.85$, lambd as $0.05$, eta as $0.001$, temperature as $5$, batch size of SimCLR loss as $64$, the number of augmentation in SimCLR loss as $2$, and the number of epochs per task as $10$.
}\end{itemize}
 
 \textbf{Combined with different adversarial training methods.}
 The learning rate, batch size, and other hyper-parameters associated with the optimization algorithm are all consistent with the ER algorithm.
 \begin{itemize}{
 \item \textbf{ER+TRADES.}
 When ER+ TRADES combines with ours, the loss of task t can be normalized as:
 \begin{equation}
 \begin{aligned}
 \label{eq:er+trades+ours}
 \mathcal{L}_{t} &\triangleq\  \mathrm{CE}(f_{\theta}({{x_t}}),y_{t})
  + \beta* \mathrm{KL}(f_{\theta}({x_t}),f_{\theta}({\widetilde{x_t})})\\
 &+\mathrm{CE}(f_{\theta}({{x_{\mathcal{M}}}}),y_{\mathcal{M}})
 + \beta* \mathrm{KL}(f_{\theta}({x_\mathcal{M}}),f_{\theta}({\widetilde{x_\mathcal{M}})}),
 \end{aligned}
 \end{equation}
where $\beta$ of TRADES is 6.0.
 \item \textbf{ER+FAT.}
 When ER+ FAT combines with ours, the loss of task t can be normalized as: 
 \begin{equation}
 \begin{aligned}
 \label{eq:er+fat}
 \mathcal{L}_{t} &\triangleq \  \mathrm{CE}(f_{\theta}({\widetilde{x_t}}),y_{t})
 +\mathrm{CE}(f_{\theta}({\widetilde{x}_{\mathcal{M}}}),y_{\mathcal{M}}).
 \end{aligned}
 \end{equation}
 Note that when solving the adversarial sample in the training phase, the iteration is stopped once the attack model is successful.

 \item \textbf{ER+LBGAT.} Here we implement LBGAT based on TRADES ($\beta=0.0$).
When ER+ LBGAT combines with ours, the loss of task t can be normalized as:
\begin{equation}
\begin{aligned}
\label{eq:er+lbgat}
\mathcal{L}_{t} &\triangleq \  \mathrm{CE}(f_{\theta}({\widetilde{x_t}}),y_{t})
 + \gamma* \mathrm{MSE}(f^{clean}_{\theta}({x_t}),f_{\theta}({\widetilde{x_t})})\\
&+\mathrm{CE}(f_{\theta}({\widetilde{x}_{\mathcal{M}}}),y_{\mathcal{M}})\\
& + \gamma* \mathrm{MSE}(f^{clean}_{\theta}({x_\mathcal{M}}),f_{\theta}({\widetilde{x}_\mathcal{M})}),
\end{aligned}
\end{equation}
$\gamma$ of LBGAT is 0.1, and $f^{clean}_{\theta}$ is a standard continual learning model (ER on our experiments) with the model architecture of ResNet-18.

\item \textbf{ER+SCORE.}
Compared with ER+TRADES, it performs better on clean samples but is less adversarial robust, probably because the hyper-parameters are unsuitable for continual learning scenarios. We implement it using $\beta$ as $4.0$, label smoothing as $0.1$, and gradient clip $g$ as $0$.
\begin{equation}
\begin{aligned}
\label{eq:er+score}
\mathcal{L}_{t} &\triangleq \operatorname{MSE}\left(f_{\theta}({x_t}),y_{t}\right)\\
& + \beta* \mathrm{ReLU}({\mathrm{MSE}(f_{\theta}({x_t}),f_{\theta}({\widetilde{x_t})})} - g)\\
&+ \operatorname{MSE}\left(f_{\theta}({x_{\mathcal{M}}}),y_{\mathcal{M}}\right)\\
& + \beta* \mathrm{ReLU}({\mathrm{MSE}(f_{\theta}({x_{\mathcal{M}}}),f_{\theta}({\widetilde{x}_{\mathcal{M}})})} - g).\\
\end{aligned}
\end{equation}

}\end{itemize}
\begin{table*}[]
    \caption{Experiment results on Split-CIFAR10/100 datasets and model architecture is ResNet18.
    Here, PGD-20 and AA are adversarial data Final Average Accuracy (FAA) generated by PGD-20 and Auto Attack (AA), respectively. Forgetting of adversarial data is computed based on PGD-20. \textbf{Bold} represents the best experimental results for the same settings, \underline{underline} indicates the second-best result, \textcolor{green}{grean} signifies relative improvement and \textcolor{red}{red} indicates relative degradation. By applying our methods, various model performances can be improved across the board.}
    \label{MainCIFAR}
\begin{minipage}{\linewidth}
    \centering
    \subcaption{Results on Split-CIFAR10. We chose two buffer sizes of $200$ and $5,120$.}
    \label{Main-CIFAR10}
    \centering 
    \resizebox{\textwidth}{!}{
    \setlength{\tabcolsep}{1pt}
    \begin{tabular}{ll l cccccccccc}
    \toprule
    \multirow{3}{*}{\textbf{Buffer Size}} &
    \multirow{3}{*}{\textbf{Method}} &\multirow{3}{*}{\textbf{Publication}}&
    \multicolumn{5}{c}{\textbf{Class Incremental Setting}} &
    \multicolumn{5}{c}{\textbf{Task Incremental Setting}} \\
    ~&~&~&\multicolumn{2}{c}{\textbf{Clean Data}}&
    \multicolumn{3}{c}{\textbf{Adversarial Data}}&
    \multicolumn{2}{c}{\textbf{Clean Data}}&
    \multicolumn{3}{c}{\textbf{Adversarial Data}}\\
    ~&~&~&\textbf{FAA$\uparrow$ }&\textbf{Forgetting$\downarrow$}&
    \textbf{PGD-20$\uparrow$}&
    \textbf{AA$\uparrow$}&
    \textbf{Forgetting$\downarrow$}&
    \textbf{FAA$\uparrow$}&
    \textbf{Forgetting$\downarrow$}&
    \textbf{PGD-20$\uparrow$}&
    \textbf{AA$\uparrow$}&
    \textbf{Forgetting$\downarrow$}\\
    \midrule
    \multirow{6}{*}{\textbf{200}}&\textbf{ER+AT}&ICLR~2018& $28.18$ & $80.58$ & $17.86$ &$16.94$ & $69.58$ & $84.49$ & $10.23$ & $44.30$ & $44.69$ & $36.89$  \\
    ~&\textbf{ER+TRADES}&ICML~2019& $22.42$& $77.25$& $15.72$& $15.53$& $64.95$& $78.79$& \underline{$8.10$}& \underline{$51.33$}& \underline{$51.50$}& $21.97$  \\
    ~&\textbf{ER+FAT}&ICML~2020&$33.61$ & $69.21$ & $15.14$ & $14.81$ & $49.04$& $83.40$ & $10.35$ & $43.69$ & $43.96$ & $28.56$  \\
 ~&\textbf{ER+LBGAT}&ICCV~2021& $25.68$& $84.47$& $16.65$& $16.56$& $70.50$& $78.19$& $18.85$& $40.69$& $40.73$& $40.83$\\
 ~&\textbf{ER+SCORE}&ICML~2022& \pmb{$48.65$} & \underline{$56.79$}& $2.40$& $0.93$& \pmb{$18.96$}& \pmb{$88.90$}& {$9.82$}& $7.25$& $6.72$& \pmb{$8.70$} \\
    \cmidrule(r){2-13}
    ~&\textbf{ER+AT+Ours}& -&
    $35.68$$(\textcolor{green}{\uparrow7.50})$& $71.18$$(\textcolor{green}{\downarrow9.40})$& \underline{$18.40$}$(\textcolor{green}{\uparrow0.54})$& \underline{$18.16$}$(\textcolor{green}{\uparrow1.22})$& $67.85$$(\textcolor{green}{\downarrow1.73})$& \underline{$84.87$}$(\textcolor{green}{\uparrow0.38})$& $9.93$$(\textcolor{green}{\downarrow0.30})$& $47.30$$(\textcolor{green}{\uparrow3.00})$& $47.61$$(\textcolor{green}{\uparrow2.92})$& $34.04$$(\textcolor{green}{\downarrow2.85})$ \\
    ~&\textbf{ER+TRADES+Ours}&-& 
    \underline{$43.34$}$(\textcolor{green}{\uparrow20.92})$
    & \pmb{$33.40$}$(\textcolor{green}{\downarrow43.85})$ 
    & \pmb{$19.85$}$(\textcolor{green}{\uparrow4.13})$
    &  \pmb{$18.35$}$(\textcolor{green}{\uparrow2.82})$
    & \underline{$30.78$}$(\textcolor{green}{\downarrow34.17})$
    & $82.59$$(\textcolor{green}{\uparrow3.80})$
    &\pmb{ $7.53$ }$(\textcolor{green}{\downarrow0.57})$
    &\pmb{ $59.41$} $(\textcolor{green}{\uparrow8.08})$
    & \pmb{$59.59$ }$(\textcolor{green}{\uparrow8.09})$
    & \underline{$14.16$}$(\textcolor{green}{\downarrow7.81})$\\
    \cmidrule(r){1-13}
    \multirow{6}{*}{\textbf{5,120}}&\textbf{ER+AT}&ICLR~2018 & $61.88$ & \underline{$37.72$} & $27.28$ & \underline{$26.69$} & $41.66$ & $91.24$ & $2.56$ & $56.59$ & $56.90$ & $19.34$\\
    ~&\textbf{ER+TRADES}&ICML~2019& $20.36$& $85.14$& $16.30$& $16.18$& $72.85$& $88.48$& \underline{$1.59$} & \underline{$64.36$}& \underline{$64.52$}& $12.47$   \\
    ~&\textbf{ER+FAT}&ICML~2020& $54.55$& $43.18$& $19.68$& $18.91$& $42.15$& \underline{$91.50$}& $2.12$& $56.72$& $56.87$& $14.79$  \\
    ~&\textbf{ER+LBGAT}&ICCV~2021& \underline{$62.45$}& $37.73$& \underline{$27.42$} & $26.66$& $47.83$& $91.10$& $3.25$& $56.57$& $56.31$& $19.38$\\
   ~&\textbf{ER+SCORE}&ICML~2022& $51.37$& $52.84$& $3.14$&$1.10$& \pmb{$20.29$}& \pmb{$95.63$}& \pmb{$1.58$}& $14.66$&$13.92$& \pmb{$2.53$}\\
    \cmidrule(r){2-13}
    ~&\textbf{ER+AT+Ours}& - &
    \pmb{$64.34$}$(\textcolor{green}{\uparrow2.46})$
    &\pmb{$23.64$}$(\textcolor{green}{\downarrow14.08})$
    &\pmb{$31.31$}$(\textcolor{green}{\uparrow4.03})$
    &\pmb{$30.49$}$(\textcolor{green}{\uparrow3.80})$ &\underline{$20.46$}$(\textcolor{green}{\downarrow21.20})$
    &$91.00$$(\textcolor{red}{\downarrow0.24})$ 
    & $3.51$$(\textcolor{red}{\uparrow0.95})$ 
    &$60.49$$(\textcolor{green}{\uparrow3.90})$
    &$60.61$ $(\textcolor{green}{\uparrow3.71})$
    &$13.27$ $(\textcolor{green}{\downarrow6.07})$\\
   ~&\textbf{ER+TRADES+Ours} & -&
   $39.80$$(\textcolor{green}{\uparrow19.44})$
   & $44.08$$(\textcolor{green}{\downarrow41.06})$
   & $23.07$$(\textcolor{green}{\uparrow6.77})$
   & $21.98$$(\textcolor{green}{\uparrow5.80})$
   & $41.37$$(\textcolor{green}{\downarrow31.48})$
   & $86.48$$(\textcolor{red}{\downarrow2.00})$
   & $1.71$$(\textcolor{red}{\uparrow0.12})$
   & \pmb{$69.25$}$(\textcolor{green}{\uparrow4.89})$ 
   & \pmb{$69.38$}$(\textcolor{green}{\uparrow4.86})$ 
   &\pmb{$5.33$}$(\textcolor{green}{\downarrow7.14})$\\
\bottomrule
\end{tabular}
}
\end{minipage}

\begin{minipage}{\linewidth}\hspace{.15in}
    \subcaption{Results on Split-CIFAR100. We choose two buffer sizes of $500$ and $2,000$.}
    \label{Main-CIFAR100}
    \centering 
    \resizebox{\textwidth}{!}{
    \setlength{\tabcolsep}{1pt}
    \begin{tabular}{ll l cccccccccc}
    \toprule
    \multirow{3}{*}{\textbf{Buffer Size}} &
    \multirow{3}{*}{\textbf{Method}} &\multirow{3}{*}{\textbf{Publication}}&
    \multicolumn{5}{c}{\textbf{Class Incremental Setting}} &
    \multicolumn{5}{c}{\textbf{Task Incremental Setting}} \\
    ~&~&~&\multicolumn{2}{c}{\textbf{Clean Data}}&
    \multicolumn{3}{c}{\textbf{Adversarial Data}}&
    \multicolumn{2}{c}{\textbf{Clean Data}}&
    \multicolumn{3}{c}{\textbf{Adversarial Data}}\\
    ~&~&~&\textbf{FAA$\uparrow$ }&\textbf{Forgetting$\downarrow$}&
    \textbf{PGD-20$\uparrow$}&
    \textbf{AA$\uparrow$}&
    \textbf{Forgetting$\downarrow$}&
    \textbf{FAA$\uparrow$}&
    \textbf{Forgetting$\downarrow$}&
    \textbf{PGD-20$\uparrow$}&
    \textbf{AA$\uparrow$}&
    \textbf{Forgetting$\downarrow$}\\
    \midrule
    \multirow{6}{*}{\textbf{500}}&\textbf{ER+AT}&ICLR~2018& $11.94$& $73.54$& $5.66$& $5.54$& $38.03$& $52.71$& $28.44$& $17.35$& $19.56$& $25.67$  \\
    ~&\textbf{ER+TRADES}&ICML~2019& $7.59$& $68.13$& $5.43$& $5.11$& $44.13$& $50.75$& \underline{$20.22$}& \underline{$26.89$}& \underline{$27.69$}& $20.34$   \\
    ~&\textbf{ER+FAT}&ICML~2020& $11.48$& $73.99$& $5.35$& $5.23$& $37.26$& $56.10$& $24.71$& $20.01$& $22.99$& $22.31$  \\
    ~&\textbf{ER+LBGAT}&ICCV~2021& $11.77$& $75.10$& $6.16$& $5.82$& $39.77$& $42.02$& $26.45$& $13.99$& $14.43$& $23.14$ \\
    ~&\textbf{ER+SCORE}&ICML~2022& $16.22$& $77.71$& $0.91$& $0.62$& \pmb{$6.13$}& \pmb{$68.87$}& \pmb{$11.21$}& $3.23$& $7.95$& \pmb{$4.94$}\\

    \cmidrule(r){2-13}
    ~&\textbf{ER+AT+Ours}& -&
    \pmb{$24.14$}$(\textcolor{green}{\uparrow12.20})$
    &\underline{$53.03$}$(\textcolor{green}{\downarrow20.51})$ 
    &\underline{$7.13$}$(\textcolor{green}{\uparrow1.47})$ 
    &\underline{$6.68$}$(\textcolor{green}{\uparrow1.14})$
    &$25.81$$(\textcolor{green}{\downarrow12.22})$
    &$56.00$$(\textcolor{green}{\uparrow3.29})$
    &$26.19$$(\textcolor{green}{\downarrow2.25})$
    &$17.95$$(\textcolor{green}{\uparrow0.60})$
    &$20.26$$(\textcolor{green}{\uparrow0.70})$
    &$24.87$$(\textcolor{green}{\downarrow0.80})$\\
    ~&\textbf{ER+TRADES+Ours}& -&
   \underline{$23.24$}$(\textcolor{green}{\uparrow15.65})$ 
   &\pmb{$24.29$}$(\textcolor{green}{\downarrow43.84})$ 
   &\pmb{$9.93$}$(\textcolor{green}{\uparrow4.50})$ 
   &\pmb{$7.50$}$(\textcolor{green}{\uparrow2.39})$ 
   &\underline{$11.70$}$(\textcolor{green}{\downarrow32.43})$ 
   &\underline{$56.68$}$(\textcolor{green}{\uparrow5.93})$ 
   &$21.39$$(\textcolor{red}{\uparrow1.17})$ 
   &\pmb{$27.39$}$(\textcolor{green}{\uparrow0.50})$ 
   &\pmb{$28.50$}$(\textcolor{green}{\uparrow0.81})$ 
   &\underline{$16.76$}$(\textcolor{green}{\downarrow3.58})$ \\

    \cmidrule(r){1-13}
    \multirow{6}{*}{\textbf{2,000}}&\textbf{ER+AT}&ICLR~2018& $18.77$& $65.06$& $7.20$& $7.01$& $33.56$& $62.01$& $17.97$& $21.16$& $24.04$& $20.47$ \\
    ~&\textbf{ER+TRADES}&ICML~2019& $9.50$& $70.49$& $5.35$& $5.01$& $42.08$& $60.63$& \underline{$13.78$}& \underline{$26.68$}& \underline{$29.19$}& $18.60$   \\
    ~&\textbf{ER+FAT}&ICML~2020& $17.09$& $66.91$& $6.12$& $5.93$& $33.62$& \underline{$63.88$}& $15.99$& $23.48$& $26.75$& $17.12$  \\
    ~&\textbf{ER+LBGAT}&ICCV~2021& $20.58$& $63.31$& $7.93$& $7.05$& $30.09$& $55.77$& $25.11$& $17.95$& $18.88$& $19.18$\\
~&\textbf{ER+SCORE}&ICML~2022& $30.10$& $61.51$& $0.75$&$0.51$& \pmb{$4.10$}& \pmb{$76.90$}& $19.60$& $4.88$&$11.97$& \pmb{$5.25$}\\
    \cmidrule(r){2-13}
    ~&\textbf{ER+AT+Ours}& - &
    \pmb{$31.93$}$(\textcolor{green}{\uparrow13.16})$
    &\underline{$40.50$}$(\textcolor{green}{\downarrow24.56})$
    &\underline{$9.61$}$(\textcolor{green}{\uparrow2.41})$
    &\underline{$9.16$}$(\textcolor{green}{\uparrow2.15})$
    &$17.78$$(\textcolor{green}{\downarrow15.78})$
    &$63.77$$(\textcolor{green}{\uparrow1.76})$
    &$17.16$$(\textcolor{green}{\downarrow0.81})$
    &$23.11$$(\textcolor{green}{\uparrow1.95})$
    &$25.59$$(\textcolor{green}{\uparrow1.55})$
    &$17.57$$(\textcolor{green}{\downarrow2.90})$\\
    ~&\textbf{ER+TRADES+Ours}& - &
    \underline{$28.73$}$(\textcolor{green}{\uparrow19.23})$
    &\pmb{$24.16$}$(\textcolor{green}{\downarrow46.33})$
    &\pmb{$12.75$}$(\textcolor{green}{\uparrow7.40})$
    &\pmb{$11.02$}$(\textcolor{green}{\uparrow6.01})$
    &\underline{$14.56$}$(\textcolor{green}{\downarrow27.52})$
    &$62.01$$(\textcolor{green}{\uparrow1.38})$
    &\pmb{$13.16$}$(\textcolor{green}{\downarrow0.62})$
    &\pmb{$34.81$}$(\textcolor{green}{\uparrow8.13})$
    &\pmb{$35.36$}$(\textcolor{green}{\uparrow6.17})$
    &\underline{$11.60$}$(\textcolor{green}{\downarrow7.00})$\\
    \bottomrule
    \end{tabular}}
\end{minipage}

\end{table*}

\textbf{Evaluation metrics.}
\label{sec:metrics}
To better measure the concerns of both adversarial robustness and continual learning, we computed Final Average Accuracy (FAA) and forgetting for the adversarial and clean samples, respectively.
Projected gradient descent (PGD) attack and Auto Attack (AA)~\cite{croce2021robustbench} are two common and effective adversarial attack methods in evaluating adversarial robustness~\cite{pang2021bag,pang2022robustness,wang2023better}.
PGD is a strong and classic white-box attack and we set its iteration as $20$ during testing.
AA is an ensemble of four diverse black-box and white-box attacks to reliably evaluate robustness, which has been proven to be reliable in evaluating deterministic defenses like adversarial training.
Additional black-box attack (RayS~\cite{chen2020rays}) evaluation results are in the Appendix.

Here, we set $a^t_i$ as the accuracy for the $i-{th}$ task after training on task t.
FAA can be defined as:
\begin{equation}
\begin{aligned}
\label{eq:faa}
\mathrm{FAA} \triangleq \frac{1}{T}\sum_{i=1}^T a^T_i,
\end{aligned}
\end{equation}
and forgetting can be defined as:
\begin{equation}
\begin{aligned}
\label{eq:forgetting}
\mathrm{forgetting} \triangleq \frac{1}{T-1}\sum_{j=1}^{T-1} f_j, \ s.t.\ f_j = \max_{l \in \{1, ..., T-1\}} a^l_i - a^{T}_j.
\end{aligned}
\end{equation}
Forgetting ranges from [-100, 100] and measures the average decrease in accuracy, \textit{i.e.}, the maximum difference in performance with respect to a given task observed over training.

Furthermore, CRD, FRI, and RRD in the analysis section can be defined as:
\begin{equation}
\begin{aligned}
\label{eq:CRD}
\mathrm{CRD} \triangleq \mathrm{FAA}_{\mathrm{clean}}-\mathrm{\widetilde{F}AA}_{\mathrm{clean}},
\end{aligned}
\end{equation}
\begin{equation}
\begin{aligned}
\label{eq:FRI}
\mathrm{FRI} \triangleq \mathrm{\widetilde{F}orgetting}_{\mathrm{clean}}-\mathrm{Forgetting}_{\mathrm{clean}},
\end{aligned}
\end{equation}
\begin{equation}
\begin{aligned}
\label{eq:RRD}
\mathrm{RRD} \triangleq (\mathrm{\widetilde{F}AA}^{\mathrm{Joint}}_{\mathrm{adv}}-\mathrm{FAA}^{\mathrm{Joint}}_{\mathrm{adv}}) -(\mathrm{\widetilde{F}AA}_{\mathrm{adv}}-\mathrm{FAA}_{\mathrm{adv}}),
\end{aligned}
\end{equation}
where $\mathrm{FAA}_{\mathrm{clean}}$ is clean data FAA of standard continual learner, $\mathrm{\widetilde{F}AA}_{\mathrm{clean}}$ is clean data FAA of adversarial robust continual learner; samely $\mathrm{FAA}_{\mathrm{adv}}$ and $\mathrm{\widetilde{F}AA}_{\mathrm{adv}}$ are adversarial data FAA of standard continual learner and adversarial robust continual learner, respectively; $\mathrm{Forgetting}_{\mathrm{clean}}$ and $\mathrm{\widetilde{F}orgetting}_{\mathrm{clean}}$ are clean data forgetting of standard continual learner and adversarial robust continual learner, respectively.
$\mathrm{\widetilde{F}AA}^{\mathrm{Joint}}_{\mathrm{adv}}$ is the adversarial FAA of the joint adversarial learner, and $\mathrm{FAA}^{\mathrm{Joint}}_{\mathrm{adv}}$ is the adversarial FAA of joint learner without adversarial training. 

\begin{figure*}[!h]
  \centering
  \includegraphics[width=1.0\linewidth]{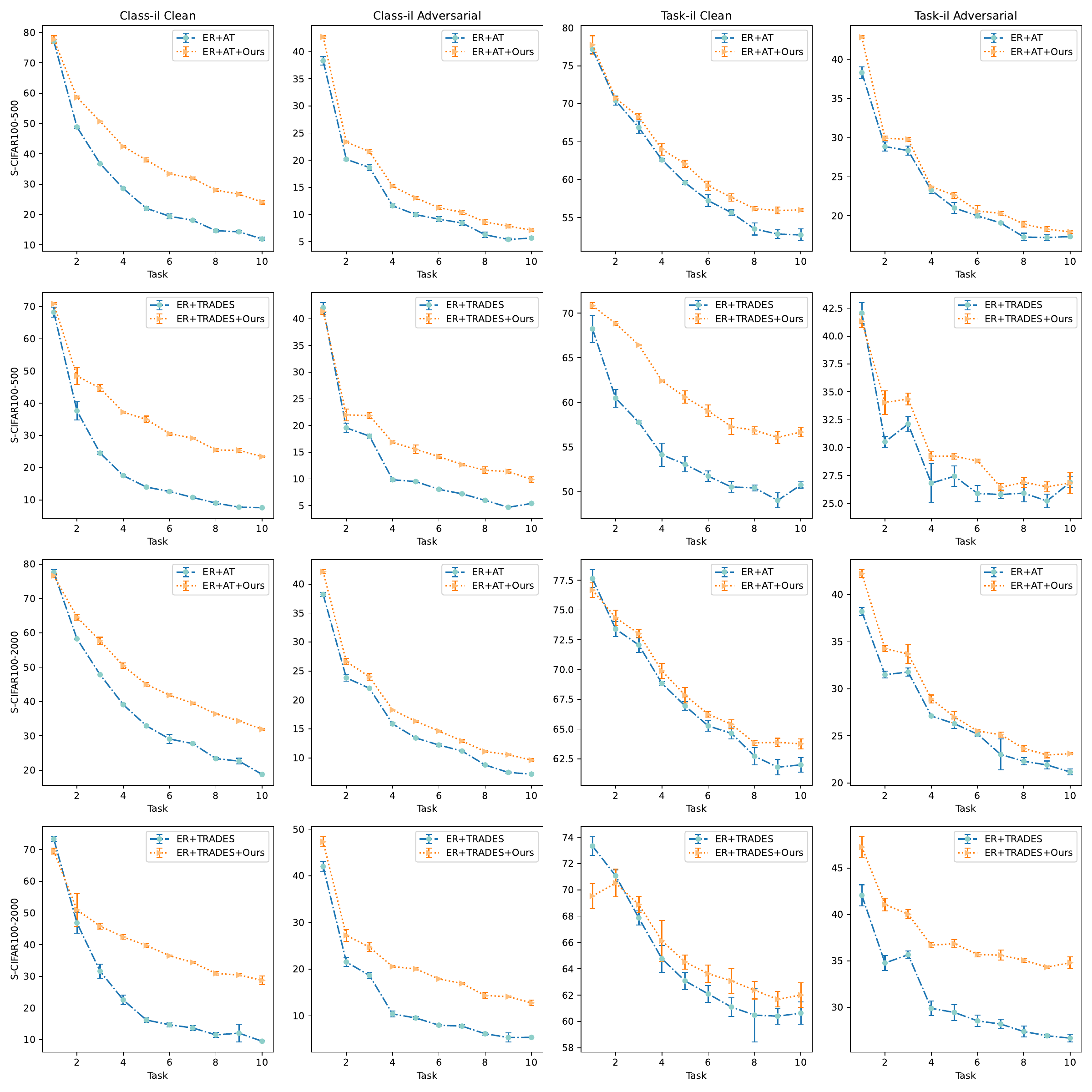}
  \caption{Accuracy curves on the Split-CIFAR100 with buffer size $500$, and $2,000$ settings.
  The plots' x-axis denotes the total number of tasks trained cumulatively up to each learning stage. The y-axis shows the average accuracy of the current task at each respective stage.
  The results demonstrate consistent improvements across most stages of continual learning when our proposed approach is combined with the baseline model.
  }
  \label{fig:task_acc_cifar100}
\end{figure*}

\subsection{Main Results}
This section compares the results of different adversarial training algorithms and memory-based continual learning methods and combines our method with them respectively on Split-CIFAR10, Split-CIFAR100, and Split-Tiny-ImageNet datasets.
Due to the high training computational cost of adversarial training, we elect to combine ER, a simple but effective method, with other adversarial training techniques to study their performances in continual learning settings.
Considering the superior performance of ER+AT and ER+TRADES ($\beta = 6.0$) in the four baselines in adversarial FAA, we combine our approach with them.
\begin{table*}[]
    \centering 
     \caption{Experiment results on Split-Tiny-ImageNet dataset and model architecture is ResNet18.
    Following~\cite{buzzega2020dark}, we choose two buffer sizes of 200 and 5,120.
    Here PGD-20 and AA are adversarial data Final Average Accuracy (FAA) generated by PGD-20 and Auto Attack (AA) respectively. Forgetting of adversarial data is computed based on PGD-20. \textbf{Bold} represents the best experimental results for the same settings, \underline{underline} indicates the second-best result, \textcolor{green}{grean} signifies relative improvement and \textcolor{red}{red} indicates relative degradation. With the addition of ours, model performance can be improved across the board.}
    \label{Main-TinyImg}
    \resizebox{\textwidth}{!}{
    \setlength{\tabcolsep}{1pt}
    \begin{tabular}{llllllllllll}
    \toprule
    \multirow{3}{*}{\textbf{Buffer Size}} &
    \multirow{3}{*}{\textbf{Method}} &
    \multicolumn{5}{c}{\textbf{Class Incremental Setting}} &
    \multicolumn{5}{c}{\textbf{Task Incremental Setting}} \\
    ~&~&\multicolumn{2}{c}{\textbf{Clean Data}}&
    \multicolumn{3}{c}{\textbf{Adversarial Data}}&
    \multicolumn{2}{c}{\textbf{Clean Data}}&
    \multicolumn{3}{c}{\textbf{Adversarial Data}}\\
    ~&~&\textbf{FAA$\uparrow$ }&\textbf{Forgetting$\downarrow$}&
    \textbf{PGD-20$\uparrow$}&
    \textbf{AA$\uparrow$}&
    \textbf{Forgetting$\downarrow$}&
    \textbf{FAA$\uparrow$}&
    \textbf{Forgetting$\downarrow$}&
    \textbf{PGD-20$\uparrow$}&
    \textbf{AA$\uparrow$}&
    \textbf{Forgetting$\downarrow$}\\
    \midrule
    \multirow{2}{*}{\textbf{200}}&\textbf{ER+TRADES} & $5.50$ & $54.67$ & $2.04$ & $1.92$ & $22.62$ & $23.51$ & \pmb{$34.72$} & $5.28$ & $6.48$ & $19.10$    \\
    ~&\textbf{ER+TRADES+Ours}
    &\pmb{$7.35$}$(\textcolor{green}{\uparrow1.85})$
    &\pmb{$52.38$}$(\textcolor{green}{\downarrow2.29})$ 
    &\pmb{$2.22$}$(\textcolor{green}{\uparrow0.18})$ 
    &\pmb{$2.05$}$(\textcolor{green}{\uparrow0.13})$
    &\pmb{$20.79$}$(\textcolor{green}{\downarrow1.83})$ 
    &\pmb{$25.58$}$(\textcolor{green}{\uparrow2.07})$
    &$34.94$$(\textcolor{red}{\uparrow0.22})$ 
    &\pmb{$6.24$}$(\textcolor{green}{\uparrow0.96})$
    &\pmb{$7.60$}$(\textcolor{green}{\uparrow1.12})$ 
    &\pmb{$18.41$}$(\textcolor{green}{\downarrow0.69})$\\
    \cmidrule(r){1-12}
    \multirow{2}{*}{\textbf{5,120}}&\textbf{ER+TRADES}& $7.24$ & $55.69$ & $2.39$ & $2.21$ & \pmb{$21.59$} & $41.36$ & $18.16$ & $11.01$ & $13.71$ & $12.61$ \\
    ~&\textbf{ER+TRADES+Ours}&\pmb{$10.95$}$(\textcolor{green}{\uparrow3.71})$
    &\pmb{$51.63$}$(\textcolor{green}{\downarrow4.06})$ 
    &\pmb{$3.01$}$(\textcolor{green}{\uparrow0.62})$ 
    &\pmb{$2.79$}$(\textcolor{green}{\uparrow0.58})$ 
    &$21.74$$(\textcolor{red}{\uparrow0.15})$ 
    &\pmb{$44.81$}$(\textcolor{green}{\uparrow3.45})$ 
    &\pmb{$15.20$}$(\textcolor{green}{\downarrow2.96})$ 
    &\pmb{$12.92$}$(\textcolor{green}{\uparrow1.91})$ 
    &\pmb{$16.21$}$(\textcolor{green}{\uparrow2.50})$
    &\pmb{$12.33$}$(\textcolor{green}{\downarrow0.28})$\\
    \bottomrule
    \end{tabular}
    }
\end{table*} 
\begin{table*}[h]
    \centering 
    \caption{Experiments with other data selection-based and logit masking-based continual learning methods. }
    \label{Data_selection_and_logit_masking}
    \resizebox{\textwidth}{!}{
    \setlength{\tabcolsep}{1pt}
    \begin{tabular}{ll cccc cccc}
    \toprule
    \multirow{3}{*}{\textbf{Method}}&\multirow{3}{*}{\textbf{Publication}}&
    \multicolumn{4}{c}{\textbf{Class Incremental Setting}} &
    \multicolumn{4}{c}{\textbf{Task Incremental Setting}} \\
    ~&~&\multicolumn{2}{c}{\textbf{Clean Data}}&
    \multicolumn{2}{c}{\textbf{Adversarial Data}}&
    \multicolumn{2}{c}{\textbf{Clean Data}}&
    \multicolumn{2}{c}{\textbf{Adversarial Data}}\\
    ~&~&\textbf{FAA$\uparrow$ }&\textbf{Forgetting$\downarrow$}&
    \textbf{FAA$\uparrow$}&
    \textbf{Forgetting$\downarrow$}&
    \textbf{FAA$\uparrow$}&\textbf{Forgetting$\downarrow$}&
    \textbf{FAA$\uparrow$}&\textbf{Forgetting$\downarrow$}\\
     \midrule
    ER+AT&\multirow{2}{*}{NeurIPS~$2019$}& $28.18$ & $80.58$ & $17.86$ & $69.58$ & $84.49$ & $10.23$ & $44.30$ &$36.89$\\
    ER+AT+Ours&~&$35.68$$(\textcolor{green}{\uparrow7.50})$& $71.18$$(\textcolor{green}{\downarrow9.40})$& {$18.40$}$(\textcolor{green}{\uparrow0.54})$& $67.85$$(\textcolor{green}{\downarrow1.73})$& {$84.87$}$(\textcolor{green}{\uparrow0.38})$& $9.93$$(\textcolor{green}{\downarrow0.30})$& $47.30$$(\textcolor{green}{\uparrow3.00})$& $34.04$$(\textcolor{green}{\downarrow2.85})$ \\   
    \midrule
    GSS+AT&\multirow{2}{*}{NeurIPS~$2019$}&{$27.59$} &{$80.78$} &{$16.67$} &{$68.53$} &{$84.41$} &{$9.83$} &{$44.25$} &$34.79$\\
    GSS+AT+Ours&~&{$36.93(\textcolor{green}{\uparrow9.34})$} &{$67.72(\textcolor{green}{\downarrow13.06})$} &{$16.84(\textcolor{green}{\uparrow0.17})$} &{$60.04(\textcolor{green}{\downarrow8.49})$} &{$85.57(\textcolor{green}{\uparrow1.16})$} &{$8.86(\textcolor{green}{\downarrow0.97})$} &{$47.11(\textcolor{green}{\uparrow2.86})$} &$31.83(\textcolor{green}{\downarrow2.97})$\\
    \midrule
    ASER+AT&\multirow{2}{*}{AAAI~$2021$}&{$18.85$} &{$87.78$} &{$14.06$} &{$65.57$} &{$73.87$} &{$19.01$} &{$30.65$} &$44.85$\\
    ASER+AT+Ours&~&{$24.45(\textcolor{green}{\uparrow5.61})$} &{$81.73(\textcolor{green}{\downarrow6.05})$} &{$14.91(\textcolor{green}{\uparrow0.85})$} &{$62.74(\textcolor{green}{\downarrow2.83})$} &{$77.70(\textcolor{green}{\uparrow3.83})$} &{$15.21(\textcolor{green}{\downarrow3.80})$} &{$34.50(\textcolor{green}{\uparrow3.85})$} &$38.55(\textcolor{green}{\downarrow6.30})$\\
    \midrule
    X-DER+AT&\multirow{2}{*}{TPAMI~$2022$}&{$34.04$} &{$25.13$} &{$16.82$} &{$27.84$} &{$80.80$} &{$4.96$} &{$60.83$} &$10.99$\\
    X-DER+AT+Ours&~&{$43.25(\textcolor{green}{\uparrow9.21})$} &{$20.77(\textcolor{green}{\downarrow4.36})$} &{$17.22(\textcolor{green}{\uparrow0.41})$} &{$18.56(\textcolor{green}{\downarrow9.28})$} &{$84.87(\textcolor{green}{\uparrow4.07})$} &{$1.74(\textcolor{green}{\downarrow3.21})$} &{$61.68(\textcolor{green}{\uparrow0.85})$} &$7.03(\textcolor{green}{\downarrow3.97})$\\
\bottomrule
\end{tabular}
}
\end{table*}




\noindent \textbf{Results on different adversarial training methods.}
Table~\ref{MainCIFAR} firstly presents the performance of combinations of different adversarial training and continual learning algorithms. We observe that the amount of data has a pronounced impact on algorithm performance, with most baselines benefiting from larger buffer sizes and showing improved overall performance. The exception is ER+TRADES on Split-CIFAR10, where at a buffer size of $5,120$, TRADES' detrimental effect on clean samples outweighs the positive impact of additional data. Second, regarding robustness, ER+AT demonstrates strong robustness (first or second place) in class incremental settings, while ER+TRADES performs best under task incremental learning, though with notably reduced clean sample accuracy due to adversarial samples. ER+LBGAT shows good robustness and clean sample accuracy under class incremental settings but does not stand out under task and class incremental settings. ER+FAT and ER+SCORE perform well on clean samples but poorly against adversarial samples, likely because their hyperparameters are not well-suited to continual settings. However, due to the immense computational cost, we lack the resources to tune each baseline finely.

Table~\ref{MainCIFAR} shows our method can significantly improve ER+TRADES and ER+AT in accuracy and forgetting performance for both clean and adversarial data on the Split-CIFAR10/100 datasets, especially in the class incremental setting.
In Table~\ref{MainCIFAR} (a), the maximum improvement in FAA is $20.92\%$ in clean samples and $8.09\%$ in adversarial samples.
What's more, the maximum improvement in forgetting is $43.85\%$ in clean samples and $34.17\%$ in adversarial samples.
In Table~\ref{MainCIFAR} (b),
clean sample accuracy is improved by up to $19.23\%$, robust accuracy by up to $8.13\%$, and forgetting is alleviated by up to $46.33\%$ for clean samples and $32.43\%$ for robust samples.
Figure~\ref{fig:task_acc_cifar100} shows FAAs of different continual training phrases of ER+AT, ER+TRADES, and their combination with us.
In the vast majority of experiments, our proposed method can improve the performance of the baseline models at each incremental training stage.
Due to length, the results of Split-CIFAR10 are detailed in the Appendix.

The results in Table~\ref{Main-TinyImg} clearly demonstrate that our proposed method is also effective at improving upon the baseline algorithms on the more challenging Split-Tiny-ImageNet dataset. Specifically, our approach led to maximum improvements in clean FAA of $3.71\%$, adversarial FAA of $2.50\%$, and alleviated forgetting by up to $4.06\%$. 

Our method achieves positive improvements in $93$ out of $100$ evaluation metrics across Table~\ref{MainCIFAR} and Table~\ref{Main-TinyImg}, demonstrating its effectiveness in mitigating forgetting while enhancing adversarial robustness. 
Despite a minor decline in FAA and forgetting on clean samples compared to baselines, especially under the task incremental setting with a buffer size of $5,120$, this is a result of the diminishing forgetting alleviation from AFLC as the buffer size increases.
Additionally, there is a trade-off between adversarial robustness and clean accuracy that needs to be considered.
The trade-off between clean accuracy and adversarial robustness is a known factor in adversarial training~\cite{zhang2019theoretically,madry2018towards,pang2022robustness,pang2020Rethinking}.
Nonetheless, our approach still enhances adversarial robustness, as demonstrated by the analysis of the interaction between RAER and AFLC.

\noindent \textbf{Results on different continual learning methods.}
We compare and combine two data selection-based continual learning methods, namely GSS and ASER, and a logit masking-based continual learning method, X-DER, using the Split-CIFAR10 dataset with a buffer size of $200$. They all only focus on continual settings without adversarial attacks.

GSS selects diverse samples based on gradients, while ASER, also based on ER, utilizes the Shapley value to identify the most helpful data for mitigating forgetting.
Unlike them, we focus specifically on adversarial robustness in the context of continual settings, selecting data based on their robustness under adversarial attacks.

Table~\ref{Data_selection_and_logit_masking} reveals that previous data selection methods employed by GSS and ASER are not suitable for adversarial settings.
When combined with adversarial training, they do not exhibit superior robustness compared to ER+AT.
Meanwhile, we can further enhance the robustness, clean data accuracy, and anti-forgetting performance both in class incremental setting and task incremental setting by combination with them.

X-DER embraces memory update and future preparation and uses logit masking, a special case of our AFLC, to reduce overweighting negative gradients of current data for past data.
While logit masking effectively alleviates forgetting, it overly suppresses the model's learning capacity. As a result, X-DER combined with AT performs worse than ER+AT in the robust FAA in class incremental setting.
However, when combined with our method, XDER+AT exhibits further improvements in terms of both final accuracy on clean and adversarial samples and mitigation of forgetting.

Results have demonstrated that by augmenting XDER+AT with our approach, we can achieve further enhancements in performance, both on adversarial and clean samples.
The consistent and sizeable gains across these metrics highlight the capability of our method to enhance robustness. Through in-depth experiments on these benchmarks, we have provided empirical evidence that validates the applicability and usefulness of our contributions when facing adversarial attacks and catastrophic forgetting.


\begin{table*}[]
    \centering 
        \caption{Ablation experiments on the Split-CIFAR10 dataset with the buffer size of $200$, using ER+TRADES as the baseline.
        \textbf{Bold} represents the best experimental results for the same settings.
        Experiments have demonstrated that AFLC mitigates clean-sample accelerated forgetting from adversarial samples, RAER mitigates gradient obfuscation (Adversarial FAA has a boost) and robust forgetting; adding FP improves the model's ability to learn new tasks (FAA has an overall increase).}
    \label{Ablation1}
    \resizebox{0.9\textwidth}{!}{
    \begin{tabular}{lll cc cccccc}
    \toprule
    \multirow{3}{*}{\textbf{AFLC}} &
    \multirow{3}{*}{\textbf{RAER}}&
    \multirow{3}{*}{\textbf{FP}}&
    \multicolumn{4}{c}{\textbf{Class Incremental Setting}} &
    \multicolumn{4}{c}{\textbf{Task Incremental Setting}} \\
    ~&~&~&\multicolumn{2}{c}{\textbf{Clean Data}}&
    \multicolumn{2}{c}{\textbf{Adversarial Data}}&
    \multicolumn{2}{c}{\textbf{Clean Data}}&
    \multicolumn{2}{c}{\textbf{Adversarial Data}}\\
    ~&~&~&\textbf{FAA$\uparrow$ }&\textbf{Forgetting$\downarrow$}&
    \textbf{FAA$\uparrow$}&
    \textbf{Forgetting$\downarrow$}&
    \textbf{FAA$\uparrow$}&\textbf{Forgetting$\downarrow$}&
    \textbf{FAA$\uparrow$}&\textbf{Forgetting$\downarrow$}\\
    \midrule
    & & &$22.42$& $77.25$& $15.72$& $64.95$& $78.79$& $8.10$& $51.33$& $21.97$ \\
    $\usym{2713}$ & & & $37.88$& $21.76$& $18.58$& $21.77$& $81.66$& $10.86$& $53.64$& $17.60$\\

   
    $\usym{2713}$&$\usym{2713}$& &$37.45$& \pmb{$10.29$}& $19.81$& \pmb{$9.61$}& $78.13$& $15.39$& {$55.71$}& \pmb{$14.13$} \\
    $\usym{2713}$&$\usym{2713}$&$\usym{2713}$& \pmb{$43.34$}& $33.40$& \pmb{$19.85$}& $30.78$& \pmb{$82.59$}& \pmb{$7.53$}& \pmb{$59.41$}& $14.16$  \\
\bottomrule
\end{tabular}}

\end{table*}
\subsection{Ablation Experiments}
\label{ex:ablation}
As shown in Table~\ref{Ablation1}, we have performed ablation experiments based on ER+TRADES under the Split-CIFAR10 dataset.
The results demonstrate that AFLC (Sec.~4.2) can effectively mitigate the increased forgetting caused by adversarial training under class incremental setting ($55.49\%$ for clean samples and $43.18\%$ for adversarial samples forgetting, with corresponding FAA improvements of $15.46\%$ and $2.86\%$, respectively).
AFLC does not show significant improvement in the task incremental setting due to excessive suppression of future task classification heads and the use of the same calibration value for classes within the same task.

RAER (Sec.~4.3) can further improve the robust accuracy of AFLC by $1.23\%$ for class incremental setting and $2.07\%$ for task incremental setting and reduce the robust forgetting by $12.16\%$ and $3.47\%$ respectively.
That proves the data selected by RAER describe the overall data distribution more accurately and effectively mitigate the gradient obfuscation phenomenon.

When considering the future prior adjustment (FP in Table~\ref{Ablation1}), we find that although the forgetting of the class incremental setting is higher, the FAA of both clean samples and adversarial samples has been significantly improved, and the forgetting of the task incremental setting has been further reduced, which proves that FP can reduce negative gradients to future classes and help learn new tasks.

\noindent \textbf{Hyper-parameter sensitivity.}
We study the sensitivity of hyper-parameters $\alpha$ in AFLC and $\rho$ in RAER on the basis of ER+TRADES with a dataset of Split-CIFAR10 and a buffer size of 200.
\begin{itemize}
    \item \textbf{Impact of $\rho$.} The results are shown in Table~\ref{Ablation2}.
The value of $\rho$ in the range of $[5,10]$ is robust and ensures the selection of safe and diverse samples for storage, but when the parameter $\rho$ is too small $(1)$, the samples selected are safe but not diverse enough to improve the robustness of the model to a limited extent.
We can also find that adding only RAER does not help the robustness of the class incremental setting, because the adversarial sample is too suppressed for the past category, and the robustness in class incremental is improved when AFLC is added (as shown in Table~\ref{Ablation1}).
    \item \textbf{Impact of $\alpha$.} The results are shown in Table~\ref{Ablation3}.
Observing the experimental results, we find that as the value of $\alpha$ increases, the negative gradient impact of adversarial examples on the classification head of previous tasks decreases, indicating a stronger ability of the model to resist forgetting. However, when $\alpha$ becomes excessively large, the model's learning capacity for the current task is heavily suppressed, resulting in a decline in the model's adversarial robustness. Therefore, we choose $\alpha = 3.5$ in other experiments.
\end{itemize}

\begin{table}[]
\centering 
\caption{Data selection strategy ablation experiments on the CIFAR-10 dataset with the buffer size of $200$, using ER+TRADES as the baseline in the class/task incremental settings. When $\rho = 11$, this is equivalent to not applying the robust data selection strategy.}
    \label{Ablation2}
    \resizebox{0.45\textwidth}{!}{
    \begin{tabular}{ccccc}
    \toprule
    \multirow{2}{*}{\textbf{$\rho$}}&
    \multicolumn{2}{c}{\textbf{Class Incremental Setting}}&
    \multicolumn{2}{c}{\textbf{Task Incremental Setting}}\\
    ~&\textbf{FAA~$\uparrow$}&
    \textbf{PGD-20$\uparrow$}&
    \textbf{FAA~$\uparrow$}&
    \textbf{PGD-20$\uparrow$}\\
    \midrule
    $11$&\pmb{$22.42$}& \pmb{$15.72$}& $78.79$& $51.33$\\
    $10$& $18.04$& $15.31$& $80.38$& \pmb{$59.88$} \\
   $5$& $18.21$& {$15.50$}& \pmb{$80.52$}& $59.50$\\
    $1$& $18.35$ & $15.06$& $79.62$& $56.92$ \\
    \bottomrule
    \end{tabular}}
    
\end{table}
    





   


\begin{table}[]
    \centering 
    \caption{Ablation experiments of different $\alpha$.}
    \label{Ablation3}
    \begin{minipage}{\linewidth}
    \centering
    \subcaption{Results on Class Incremental Setting.}
    \centering 
  {
    \begin{tabular}{llcccc}
    \toprule
    \multirow{2}{*}{\textbf{$\alpha$}}&
    \multicolumn{2}{c}{\textbf{Clean Data}}&
    \multicolumn{2}{c}{\textbf{Adversarial Data}}\\
    ~&\textbf{FAA$\uparrow$ }&\textbf{Forgetting$\downarrow$}&
    \textbf{FAA$\uparrow$}&
    \textbf{Forgetting$\downarrow$}\\
    \midrule

    $0.0$& \pmb{$38.85$}& {$47.53$}& {$17.56$}& {$42.80$}\\
    $3.5$& $37.88$& $21.76$&  \pmb{$18.58$}& $21.77$\\
    $7.0$& $35.92$& \pmb{$20.99$}& $13.89$&  \pmb{$21.76$}\\
    \bottomrule
    \end{tabular}}
    \end{minipage}

    \begin{minipage}{\linewidth}\hspace{.15in}
    \centering
    \subcaption{Results on Task Incremental Setting.}
    \centering 
  {
    \begin{tabular}{llcccc}
    \toprule
    \multirow{2}{*}{\textbf{$\alpha$}}&
    \multicolumn{2}{c}{\textbf{Clean Data}}&
    \multicolumn{2}{c}{\textbf{Adversarial Data}}\\
    ~&\textbf{FAA$\uparrow$ }&\textbf{Forgetting$\downarrow$}&
    \textbf{FAA$\uparrow$}&
    \textbf{Forgetting$\downarrow$}\\
    \midrule

    $0.0$&  \pmb{$82.70$}& \pmb{$9.97$} & $50.67$& $26.28$ \\
    $3.5$&$81.66$& $10.86$& \pmb{$53.64$}& $17.60$\\
    $7.0$& $82.56$& $10.49$& $52.69$& \pmb{$16.25$}\\
\bottomrule
\end{tabular}}
\end{minipage}
\end{table}
\subsection{Generalization on Adaptive Attacks}
   

\begin{table}[]
 \caption{Adaptive Attack for ER+TRADES+Ours. All the attack methods in the table incorporate the same logit calibration as in the training phase of our model. Forgetting is based on PGD-20. Results show our method still maintains adversarial robustness under adaptive attack.}
    \label{adaptive}
    \centering 
{
    \begin{tabular}{ll cccc}
    \toprule
    \multicolumn{2}{c}{\textbf{Dataset}} &  \multicolumn{2}{c}{\textbf{S-CIFAR10}}&  \multicolumn{2}{c}{\textbf{S-CIFAR100}}\\
    \midrule
    \multicolumn{2}{c}{\textbf{Buffer Size}}&$200$&$5,120$&$500$&$2,000$\\
    \midrule
    \multirow{3}{*}{\textbf{Class-IL}} &\textbf{PGD-20}&$23.99$&$27.50$&$9.91$&$13.68$\\
    \cmidrule(r){2-6}
    ~&\textbf{AA}&$41.76$&$35.29$&$20.61$&$25.85$\\
    \cmidrule(r){2-6}
    ~&\textbf{Forgetting}&$26.43$&$1.53$&$13.10$&$15.33$\\
    \midrule
    \multirow{3}{*}{\textbf{Task-IL}} &\textbf{PGD-20}&$58.18$&$65.59$&$26.84$&$34.83$\\
    \cmidrule(r){2-6}
    ~&\textbf{AA}&$58.32$&$60.69$&$28.20$&$35.72$\\
    \cmidrule(r){2-6}
    ~&\textbf{Forgetting}&$15.40$&$5.89$&$17.13$&$11.59$\\
\bottomrule
\end{tabular}}
   
\end{table}

As mentioned by~\cite{athalye2018obfuscated, NEURIPS2020_5898d809}, generic attack methods alone are not adequate to account for solid robustness.
Therefore, we use the adaptive attack based on PGD-20 and Auto Attack by doing the same logit calibration as our model training phase when generating the adversarial samples (details in supplemental materials).
As shown in Table~\ref{adaptive}, we are still able to maintain stable robustness even under adaptive attacks.
We find that adding logit calibration to the solution adversarial sample stage reduces the attack strength, especially for AA, and Yang~\etal~\cite{yang2022a} also find a similar phenomenon.
We conjecture that logit calibration may introduce overfitting from logit prior when generating adversarial examples.

For both PGD-20 and AA, we considered both class incremental setting and task incremental setting, and the logit in solving the adversarial sample is processed by AFLC.
\begin{equation}
\begin{aligned}
\label{eq:logit-adjust-aa}
h^{\mathrm{lc}}_{\theta}(\widetilde{x})_{i} = h_{\theta}(\widetilde{x})_{i} - \mathrm{v}_i,
\end{aligned}
\end{equation}
where $\mathrm{v}_i$ is the same as the $\mathrm{v}$ of the last task training phase.




\section{Conclusion and Future Work}
In this paper, we delve into how to attain an adversarial robust continual model without additional data.
We separately select representative approaches in continual learning and adversarial defenses: memory-based continual learning and adversarial training and conduct systematic experimental and theoretical analyses of their direct combination, which reveal the inherent difficulties: accelerated forgetting for continual learners and gradient obfuscation for adversarial robustness.
We make a preliminary attempt to alleviate these two major difficulties.
(1) Anti-forgettable logit calibration is proposed to reduce the negative influences of past knowledge by correcting the logits of different classes.
(2) We design a robust difficulty-based data selection mechanism to mitigate gradient obfuscation by selecting adversarial safe and diverse data to store.
Experiments demonstrate our effectiveness, which we believe represents a step toward achieving an ideal model.

\textbf{Future work and limitations}
In this paper, we propose novel algorithms aimed at enhancing the adversarial robustness of memory-based continual learning methods.
In the future, we plan to investigate how to improve the adversarial robustness of other types of continual learning algorithms, such as dynamic architecture or regularization-based continual methods.

Nevertheless, it is important to acknowledge certain limitations of our current work. While we have achieved notable advancements in enhancing adversarial robustness within the memory-based continual learning framework, the efficacy of these algorithms might differ when applied to distinct learning architectures. Moreover, the complexities associated with real-world datasets and dynamic environments pose challenges that warrant further investigation. As we extend our methodologies to diverse settings, careful consideration of these intricacies will be imperative to ensure the robustness and practical applicability of our proposed approaches.

%

\ifCLASSOPTIONcaptionsoff
  \newpage
\fi

{\small
\bibliographystyle{IEEEtran}
\bibliography{egbib}

\begin{thebibliography}{10}
\providecommand{\url}[1]{#1}
\csname url@samestyle\endcsname
\providecommand{\newblock}{\relax}
\providecommand{\bibinfo}[2]{#2}
\providecommand{\BIBentrySTDinterwordspacing}{\spaceskip=0pt\relax}
\providecommand{\BIBentryALTinterwordstretchfactor}{4}
\providecommand{\BIBentryALTinterwordspacing}{\spaceskip=\fontdimen2\font plus
\BIBentryALTinterwordstretchfactor\fontdimen3\font minus
  \fontdimen4\font\relax}
\providecommand{\BIBforeignlanguage}[2]{{%
\expandafter\ifx\csname l@#1\endcsname\relax
\typeout{** WARNING: IEEEtran.bst: No hyphenation pattern has been}%
\typeout{** loaded for the language `#1'. Using the pattern for}%
\typeout{** the default language instead.}%
\else
\language=\csname l@#1\endcsname
\fi
#2}}
\providecommand{\BIBdecl}{\relax}
\BIBdecl

\bibitem{parisi2019continual}
G.~I. Parisi, R.~Kemker, J.~L. Part, C.~Kanan, and S.~Wermter, ``Continual
  lifelong learning with neural networks: A review,'' \emph{Neural Networks},
  vol. 113, pp. 54--71, 2019.

\bibitem{sarfraz2023error}
\BIBentryALTinterwordspacing
F.~Sarfraz, E.~Arani, and B.~Zonooz, ``Error sensitivity modulation based
  experience replay: Mitigating abrupt representation drift in continual
  learning,'' in \emph{The Eleventh International Conference on Learning
  Representations}, 2023. [Online]. Available:
  \url{https://openreview.net/forum?id=zlbci7019Z3}
\BIBentrySTDinterwordspacing

\bibitem{wu2021pretrained}
T.~Wu, M.~Caccia, Z.~Li, Y.-F. Li, G.~Qi, and G.~Haffari, ``Pretrained language
  model in continual learning: A comparative study,'' in \emph{International
  Conference on Learning Representations}, 2021.

\bibitem{yoon2019scalable}
J.~Yoon, S.~Kim, E.~Yang, and S.~J. Hwang, ``Scalable and order-robust
  continual learning with additive parameter decomposition,'' \emph{arXiv
  preprint arXiv:1902.09432}, 2019.

\bibitem{khan2022adversarially}
H.~Khan, N.~C. Bouaynaya, and G.~Rasool, ``Adversarially robust continual
  learning,'' in \emph{2022 International Joint Conference on Neural
  Networks}.\hskip 1em plus 0.5em minus 0.4em\relax IEEE, 2022, pp. 1--8.

\bibitem{chen2022queried}
T.~Chen, S.~Liu, S.~Chang, L.~Amini, and Z.~Wang, ``Queried unlabeled data
  improves and robustifies class-incremental learning,'' \emph{Transactions on
  Machine Learning and Data Mining}, 2022.

\bibitem{li2017learning}
Z.~Li and D.~Hoiem, ``Learning without forgetting,'' \emph{IEEE transactions on
  pattern analysis and machine intelligence}, vol.~40, no.~12, pp. 2935--2947,
  2017.

\bibitem{zhang2019theoretically}
H.~Zhang, Y.~Yu, J.~Jiao, E.~Xing, L.~El~Ghaoui, and M.~Jordan, ``Theoretically
  principled trade-off between robustness and accuracy,'' in
  \emph{International Conference on Machine Learning}.\hskip 1em plus 0.5em
  minus 0.4em\relax PMLR, 2019, pp. 7472--7482.

\bibitem{riemer2019learning}
M.~Riemer, I.~Cases, R.~Ajemian, M.~Liu, I.~Rish, Y.~Tu, and G.~Tesauro,
  ``Learning to learn without forgetting by maximizing transfer and minimizing
  interference,'' in \emph{International Conference on Learning
  Representations}, 2019.

\bibitem{guoattacking}
Y.~Guo, M.~Liu, Y.~Li, L.~Wang, T.~Yang, and T.~Rosing, ``Attacking lifelong
  learning models with gradient reversion,'' 2020.

\bibitem{khan2022susceptibility}
H.~Khan, P.~M. Shah, S.~F.~A. Zaidi \emph{et~al.}, ``Susceptibility of
  continual learning against adversarial attacks,'' \emph{arXiv preprint
  arXiv:2207.05225}, 2022.

\bibitem{athalye2018obfuscated}
A.~Athalye, N.~Carlini, and D.~Wagner, ``Obfuscated gradients give a false
  sense of security: Circumventing defenses to adversarial examples,'' in
  \emph{International Conference on Machine Learning}.\hskip 1em plus 0.5em
  minus 0.4em\relax PMLR, 2018, pp. 274--283.

\bibitem{chaudhry2018riemannian}
A.~Chaudhry, P.~K. Dokania, T.~Ajanthan, and P.~H. Torr, ``Riemannian walk for
  incremental learning: Understanding forgetting and intransigence,'' in
  \emph{European Conference on Computer Vision}.\hskip 1em plus 0.5em minus
  0.4em\relax Springer, 2018, pp. 532--547.

\bibitem{mccloskey1989catastrophic}
M.~McCloskey and N.~J. Cohen, ``Catastrophic interference in connectionist
  networks: The sequential learning problem,'' in \emph{Psychology of learning
  and motivation}.\hskip 1em plus 0.5em minus 0.4em\relax Elsevier, 1989,
  vol.~24, pp. 109--165.

\bibitem{ratcliff1990connectionist}
R.~Ratcliff, ``Connectionist models of recognition memory: constraints imposed
  by learning and forgetting functions.'' \emph{Psychological review}, vol.~97,
  no.~2, p. 285, 1990.

\bibitem{buzzega2020dark}
P.~Buzzega, M.~Boschini, A.~Porrello, D.~Abati, and S.~Calderara, ``Dark
  experience for general continual learning: a strong, simple baseline,''
  \emph{Advances in neural information processing systems}, vol.~33, pp.
  15\,920--15\,930, 2020.

\bibitem{caccia2022new}
L.~Caccia, R.~Aljundi, N.~Asadi, T.~Tuytelaars, J.~Pineau, and E.~Belilovsky,
  ``New insights on reducing abrupt representation change in online continual
  learning,'' in \emph{International Conference on Learning Representations},
  2022.

\bibitem{kirkpatrick2017overcoming}
J.~Kirkpatrick, R.~Pascanu, N.~Rabinowitz, J.~Veness, G.~Desjardins, A.~A.
  Rusu, K.~Milan, J.~Quan, T.~Ramalho, A.~Grabska-Barwinska \emph{et~al.},
  ``Overcoming catastrophic forgetting in neural networks,'' \emph{Proceedings
  of the national academy of sciences}, vol. 114, no.~13, pp. 3521--3526, 2017.

\bibitem{lin2022towards}
G.~Lin, H.~Chu, and H.~Lai, ``Towards better plasticity-stability trade-off in
  incremental learning: A simple linear connector,'' in \emph{Proceedings of
  the IEEE/CVF Conference on Computer Vision and Pattern Recognition}, 2022,
  pp. 89--98.

\bibitem{douillard2022dytox}
A.~Douillard, A.~Ram{\'e}, G.~Couairon, and M.~Cord, ``Dytox: Transformers for
  continual learning with dynamic token expansion,'' in \emph{Proceedings of
  the IEEE/CVF Conference on Computer Vision and Pattern Recognition}, 2022,
  pp. 9285--9295.

\bibitem{rusu2016progressive}
A.~A. Rusu, N.~C. Rabinowitz, G.~Desjardins, H.~Soyer, J.~Kirkpatrick,
  K.~Kavukcuoglu, R.~Pascanu, and R.~Hadsell, ``Progressive neural networks,''
  \emph{arXiv preprint arXiv:1606.04671}, 2016.

\bibitem{wang2022learning}
Z.~Wang, Z.~Zhang, C.-Y. Lee, H.~Zhang, R.~Sun, X.~Ren, G.~Su, V.~Perot, J.~Dy,
  and T.~Pfister, ``Learning to prompt for continual learning,'' in
  \emph{Proceedings of the IEEE/CVF Conference on Computer Vision and Pattern
  Recognition}, 2022, pp. 139--149.

\bibitem{9477031}
J.~Xu, J.~Ma, X.~Gao, and Z.~Zhu, ``Adaptive progressive continual learning,''
  \emph{IEEE Transactions on Pattern Analysis and Machine Intelligence},
  vol.~44, no.~10, pp. 6715--6728, 2022.

\bibitem{9915459}
M.~Masana, X.~Liu, B.~Twardowski, M.~Menta, A.~D. Bagdanov, and J.~van~de
  Weijer, ``Class-incremental learning: Survey and performance evaluation on
  image classification,'' \emph{IEEE Transactions on Pattern Analysis and
  Machine Intelligence}, vol.~45, no.~5, pp. 5513--5533, 2023.

\bibitem{zhou2023deep}
D.-W. Zhou, Q.-W. Wang, Z.-H. Qi, H.-J. Ye, D.-C. Zhan, and Z.~Liu, ``Deep
  class-incremental learning: A survey,'' \emph{arXiv preprint
  arXiv:2302.03648}, 2023.

\bibitem{farquhar2018towards}
S.~Farquhar and Y.~Gal, ``Towards robust evaluations of continual learning,''
  \emph{arXiv preprint arXiv:1805.09733}, 2018.

\bibitem{umer2020targeted}
M.~Umer, G.~Dawson, and R.~Polikar, ``Targeted forgetting and false memory
  formation in continual learners through adversarial backdoor attacks,'' in
  \emph{2020 International Joint Conference on Neural Networks}.\hskip 1em plus
  0.5em minus 0.4em\relax IEEE, 2020, pp. 1--8.

\bibitem{hassanpour2022differential}
A.~Hassanpour, M.~Moradikia, B.~Yang, A.~Abdelhadi, C.~Busch, and J.~Fierrez,
  ``Differential privacy preservation in robust continual learning,''
  \emph{IEEE Access}, vol.~10, pp. 24\,273--24\,287, 2022.

\bibitem{wang2022improving}
Z.~Wang, L.~Shen, L.~Fang, Q.~Suo, T.~Duan, and M.~Gao, ``Improving task-free
  continual learning by distributionally robust memory evolution,'' in
  \emph{International Conference on Machine Learning}.\hskip 1em plus 0.5em
  minus 0.4em\relax PMLR, 2022, pp. 22\,985--22\,998.

\bibitem{kumari2022retrospective}
\BIBentryALTinterwordspacing
L.~Kumari, S.~Wang, T.~Zhou, and J.~Bilmes, ``Retrospective adversarial replay
  for continual learning,'' in \emph{Advances in Neural Information Processing
  Systems}, A.~H. Oh, A.~Agarwal, D.~Belgrave, and K.~Cho, Eds., 2022.
  [Online]. Available: \url{https://openreview.net/forum?id=XEoih0EwCwL}
\BIBentrySTDinterwordspacing

\bibitem{szegedy2013intriguing}
C.~Szegedy, W.~Zaremba, I.~Sutskever, J.~Bruna, D.~Erhan, I.~Goodfellow, and
  R.~Fergus, ``Intriguing properties of neural networks,'' \emph{arXiv preprint
  arXiv:1312.6199}, 2013.

\bibitem{duan2021advdrop}
R.~Duan, Y.~Chen, D.~Niu, Y.~Yang, A.~K. Qin, and Y.~He, ``Advdrop: Adversarial
  attack to dnns by dropping information,'' in \emph{Proceedings of the
  IEEE/CVF International Conference on Computer Vision}, 2021, pp. 7506--7515.

\bibitem{zhang2022towards}
J.~Zhang, B.~Li, J.~Xu, S.~Wu, S.~Ding, L.~Zhang, and C.~Wu, ``Towards
  efficient data free black-box adversarial attack,'' in \emph{Proceedings of
  the IEEE/CVF Conference on Computer Vision and Pattern Recognition}, 2022,
  pp. 15\,115--15\,125.

\bibitem{akhtar2018threat}
N.~Akhtar and A.~Mian, ``Threat of adversarial attacks on deep learning in
  computer vision: A survey,'' \emph{IEEE Access}, vol.~6, pp.
  14\,410--14\,430, 2018.

\bibitem{bai2021recent}
T.~Bai, J.~Luo, J.~Zhao, B.~Wen, and Q.~Wang, ``Recent advances in adversarial
  training for adversarial robustness,'' \emph{arXiv preprint
  arXiv:2102.01356}, 2021.

\bibitem{croce2021robustbench}
F.~Croce, M.~Andriushchenko, V.~Sehwag, E.~Debenedetti, N.~Flammarion,
  M.~Chiang, P.~Mittal, and M.~Hein, ``Robustbench: a standardized adversarial
  robustness benchmark,'' in \emph{Thirty-fifth Conference on Neural
  Information Processing Systems Datasets and Benchmarks Track (Round 2)},
  2021.

\bibitem{cui2021learnable}
J.~Cui, S.~Liu, L.~Wang, and J.~Jia, ``Learnable boundary guided adversarial
  training,'' in \emph{Proceedings of the IEEE/CVF International Conference on
  Computer Vision}, 2021, pp. 15\,721--15\,730.

\bibitem{madry2018towards}
A.~Madry, A.~Makelov, L.~Schmidt, D.~Tsipras, and A.~Vladu, ``Towards deep
  learning models resistant to adversarial attacks,'' in \emph{International
  Conference on Learning Representations}, 2018.

\bibitem{maini2020adversarial}
P.~Maini, E.~Wong, and Z.~Kolter, ``Adversarial robustness against the union of
  multiple perturbation models,'' in \emph{International Conference on Machine
  Learning}.\hskip 1em plus 0.5em minus 0.4em\relax PMLR, 2020, pp. 6640--6650.

\bibitem{pang2022robustness}
T.~Pang, M.~Lin, X.~Yang, J.~Zhu, and S.~Yan, ``Robustness and accuracy could
  be reconcilable by (proper) definition,'' \emph{International Conference on
  Machine Learning}, 2022.

\bibitem{pang2021bag}
T.~Pang, X.~Yang, Y.~Dong, H.~Su, and J.~Zhu, ``Bag of tricks for adversarial
  training,'' in \emph{International Conference on Learning Representations},
  2021.

\bibitem{9761760}
S.~Lee, H.~Kim, and J.~Lee, ``Graddiv: Adversarial robustness of randomized
  neural networks via gradient diversity regularization,'' \emph{IEEE
  Transactions on Pattern Analysis and Machine Intelligence}, vol.~45, no.~2,
  pp. 2645--2651, 2023.

\bibitem{wu2021adversarial}
T.~Wu, Z.~Liu, Q.~Huang, Y.~Wang, and D.~Lin, ``Adversarial robustness under
  long-tailed distribution,'' in \emph{Proceedings of the IEEE/CVF conference
  on computer vision and pattern recognition}, 2021, pp. 8659--8668.

\bibitem{shao2020open}
R.~Shao, P.~Perera, P.~C. Yuen, and V.~M. Patel, ``Open-set adversarial
  defense,'' in \emph{European Conference on Computer Vision}.\hskip 1em plus
  0.5em minus 0.4em\relax Springer, 2020, pp. 682--698.

\bibitem{chou2022continual}
T.-C. Chou, J.-Y. Huang, and W.-P. Lee, ``Continual learning with adversarial
  training to enhance robustness of image recognition models,'' in \emph{2022
  International Conference on Cyberworlds}.\hskip 1em plus 0.5em minus
  0.4em\relax IEEE, 2022, pp. 236--242.

\bibitem{rostami2021detection}
M.~Rostami, L.~Spinoulas, M.~Hussein, J.~Mathai, and W.~Abd-Almageed,
  ``Detection and continual learning of novel face presentation attacks,'' in
  \emph{Proceedings of the IEEE/CVF international conference on computer
  vision}, 2021, pp. 14\,851--14\,860.

\bibitem{boschini2022class}
M.~Boschini, L.~Bonicelli, P.~Buzzega, A.~Porrello, and S.~Calderara,
  ``Class-incremental continual learning into the extended der-verse,''
  \emph{IEEE Transactions on Pattern Analysis and Machine Intelligence}, 2022.

\bibitem{krizhevsky2009learning}
\BIBentryALTinterwordspacing
A.~Krizhevsky and G.~Hinton, ``Learning multiple layers of features from tiny
  images,'' University of Toronto, Toronto, Ontario, Tech. Rep.~0, 2009.
  [Online]. Available:
  \url{https://www.cs.toronto.edu/~kriz/learning-features-2009-TR.pdf}
\BIBentrySTDinterwordspacing

\bibitem{menon2021longtail}
\BIBentryALTinterwordspacing
A.~K. Menon, S.~Jayasumana, A.~S. Rawat, H.~Jain, A.~Veit, and S.~Kumar,
  ``Long-tail learning via logit adjustment,'' in \emph{International
  Conference on Learning Representations}, 2021. [Online]. Available:
  \url{https://openreview.net/forum?id=37nvvqkCo5}
\BIBentrySTDinterwordspacing

\bibitem{zhao2022adaptive}
Y.~Zhao, W.~Chen, X.~Tan, K.~Huang, and J.~Zhu, ``Adaptive logit adjustment
  loss for long-tailed visual recognition,'' in \emph{Proceedings of the AAAI
  Conference on Artificial Intelligence}, vol.~36, no.~3, 2022, pp. 3472--3480.

\bibitem{zhang2021_GAIRAT}
\BIBentryALTinterwordspacing
J.~Zhang, J.~Zhu, G.~Niu, B.~Han, M.~Sugiyama, and M.~Kankanhalli,
  ``Geometry-aware instance-reweighted adversarial training,'' in
  \emph{International Conference on Learning Representations}, 2021. [Online].
  Available: \url{https://openreview.net/forum?id=iAX0l6Cz8ub}
\BIBentrySTDinterwordspacing

\bibitem{aljundi2019gradient}
R.~Aljundi, M.~Lin, B.~Goujaud, and Y.~Bengio, ``Gradient based sample
  selection for online continual learning,'' \emph{Advances in neural
  information processing systems}, vol.~32, 2019.

\bibitem{shim2021online}
D.~Shim, Z.~Mai, J.~Jeong, S.~Sanner, H.~Kim, and J.~Jang, ``Online
  class-incremental continual learning with adversarial shapley value,'' in
  \emph{Proceedings of the AAAI Conference on Artificial Intelligence},
  vol.~35, no.~11, 2021, pp. 9630--9638.

\bibitem{zhu2023improving}
K.~Zhu, J.~Wang, X.~Hu, X.~Xie, and G.~Yang, ``Improving generalization of
  adversarial training via robust critical fine-tuning,'' \emph{arXiv preprint
  arXiv:2308.02533}, 2023.

\bibitem{stanford2015tinyimagenet}
Stanford, ``Tiny imagenet challenge (cs231n),''
  \url{http://tiny-imagenet.herokuapp.com/}, 2015, accessed: Feb 21, 2023.

\bibitem{zhang2020attacks}
J.~Zhang, X.~Xu, B.~Han, G.~Niu, L.~Cui, M.~Sugiyama, and M.~Kankanhalli,
  ``Attacks which do not kill training make adversarial learning stronger,'' in
  \emph{International Conference on Machine Learning}.\hskip 1em plus 0.5em
  minus 0.4em\relax PMLR, 2020, pp. 11\,278--11\,287.

\bibitem{wang2023better}
Z.~Wang, T.~Pang, C.~Du, M.~Lin, W.~Liu, and S.~Yan, ``Better diffusion models
  further improve adversarial training,'' in \emph{International Conference on
  Machine Learning}, 2023.

\bibitem{chen2020rays}
J.~Chen and Q.~Gu, ``Rays: A ray searching method for hard-label adversarial
  attack,'' in \emph{Proceedings of the 26rd ACM SIGKDD International
  Conference on Knowledge Discovery and Data Mining}, 2020.

\bibitem{pang2020Rethinking}
\BIBentryALTinterwordspacing
T.~Pang, K.~Xu, Y.~Dong, C.~Du, N.~Chen, and J.~Zhu, ``Rethinking softmax
  cross-entropy loss for adversarial robustness,'' in \emph{International
  Conference on Learning Representations}, 2020. [Online]. Available:
  \url{https://openreview.net/forum?id=Byg9A24tvB}
\BIBentrySTDinterwordspacing

\bibitem{NEURIPS2020_5898d809}
\BIBentryALTinterwordspacing
T.~Pang, X.~Yang, Y.~Dong, K.~Xu, J.~Zhu, and H.~Su, ``Boosting adversarial
  training with hypersphere embedding,'' in \emph{Advances in Neural
  Information Processing Systems}, H.~Larochelle, M.~Ranzato, R.~Hadsell,
  M.~Balcan, and H.~Lin, Eds., vol.~33.\hskip 1em plus 0.5em minus 0.4em\relax
  Curran Associates, Inc., 2020, pp. 7779--7792. [Online]. Available:
  \url{https://proceedings.neurips.cc/paper/2020/file/5898d8095428ee310bf7fa3da1864ff7-Paper.pdf}
\BIBentrySTDinterwordspacing

\bibitem{yang2022a}
\BIBentryALTinterwordspacing
Z.~Yang, T.~Pang, and Y.~Liu, ``A closer look at the adversarial robustness of
  deep equilibrium models,'' in \emph{Advances in Neural Information Processing
  Systems}, A.~H. Oh, A.~Agarwal, D.~Belgrave, and K.~Cho, Eds., 2022.
  [Online]. Available: \url{https://openreview.net/forum?id=_WHs1ruFKTD}
\BIBentrySTDinterwordspacing

\end{thebibliography}
}

\end{document}